\documentclass[journal]{IEEEtran}
\usepackage{amsthm, amssymb, amsfonts,algorithm, algpseudocode, array}
\usepackage[cmex10]{amsmath}
\usepackage[english]{babel}
\usepackage{cancel,cite,color,comment,dblfloatfix}
\usepackage{enumerate,esint}
\usepackage{float}
\usepackage[T1]{fontenc}
\usepackage{graphicx}
\usepackage[latin9]{inputenc}
\usepackage{multirow,multicol}
\usepackage{slashbox,soul}
\usepackage{tikz,tgtermes}
\usepackage[labelformat=simple]{subcaption}
\usepackage{pgfplots}

\usetikzlibrary{backgrounds}
\usetikzlibrary{shapes, snakes}
\pgfplotsset{compat=1.11}%, every axis/.append style = {line width = 3pt}}
\allowdisplaybreaks
\newcommand{\dS}{{d_{p}^{(c, \gamma)}}}
\newcommand{\dSh}{{\bar{d}_{p}^{(c, \gamma)}}}

\newcommand{\dbase}{{d_{b}}}
\newcommand{\nn}{{\nonumber}}
\newcommand{\RN}{\mathbb{R}^N}
\newcommand{\Z}{\mathbb{N}}
\newcommand{\E}{\mathbb{E}}
\newcommand{\R}{\mathbb{R}}
\newcommand{\bX}{{\bf{X}}}
\newcommand{\bY}{{\bf{Y}}}
\newcommand{\bZ}{{\bf{Z}}}
\newcommand{\nx}{{n_{\bX}}}
\newcommand{\ny}{{n_{\bY}}}
\newcommand{\nz}{{n_{\bZ}}}
\newcommand{\Tk}{{\tau^{k}}}
\newcommand{\tr}{{\mathrm{tr}}}

\newcommand{\W}{{\mathcal{W}}}
\newcommand{\hW}{{\overline{\mathcal{W}}}}

\theoremstyle{plain}
\newtheorem{thm}{Theorem}
\newtheorem{lem}[thm]{Lemma}
\newtheorem{prop}[thm]{Proposition}

\theoremstyle{definition}
\newtheorem{defn}{Definition}

\theoremstyle{remark}

\begin{document}

\title{A metric on the space of finite sets of trajectories for evaluation of multi-target tracking algorithms}
\author{Ángel F. García-Fernández, Abu Sajana Rahmathullah, Lennart
	Svensson \thanks{Á. F. García-Fernández is with the Department of Electrical Engineering and Electronics, University of Liverpool, Liverpool L69 3GJ, United Kingdom, and also with the ARIES research center, Universidad Antonio de Nebrija, Madrid, Spain (email: angel.garcia-fernandez@liverpool.ac.uk). A. S. Rahmathullah is with Zenuity AB, Gothenburg, Sweden (email: abusajana@gmail.com). L. Svensson is with the Department of Signals and Systems, Chalmers University of Technology, SE-412 96 Gothenburg, Sweden (email: lennart.svensson@chalmers.se).}}
\date{\today}

\maketitle
\begin{abstract}
In this paper, we propose a metric on the space of finite sets of trajectories for assessing multi-target tracking algorithms in a mathematically sound way. The main use of the metric is to compare estimates of trajectories from different algorithms with the ground truth of trajectories. The proposed metric includes intuitive costs associated to localization error for properly detected targets, missed and false targets and track switches at each time step. The metric computation is based on solving a multi-dimensional assignment problem. We also propose a lower bound for the metric, which is also a metric for sets of trajectories and is computable in polynomial time using linear programming. We also extend the proposed metrics on sets of trajectories to random finite sets of trajectories.
\end{abstract}

\section{Introduction} \label{sec:intro}

The main goal of multiple target tracking (MTT) is to estimate a collection of trajectories, which represent the evolution of target states over time, from noisy sensor observations \cite{mahler2007statistical}. To evaluate the quality of estimates provided by different algorithms, one needs a distance function that quantifies the error between the ground truth, which represents the true trajectories, and the estimate. In order to design such a distance function, first we need a space, where both the ground truth and the estimate lie. In the typical MTT models, targets are born, move and die \cite{mahler2007statistical}. Therefore, a natural and minimal representation of the ground truth and its estimates is a set of trajectories \cite{garcia2015setsTraj}, where a trajectory is a sequence of target states with a time of appearance and a certain length. Second, it is desirable that the distance function\footnote{The terms metric and distance are often used interchangeably \cite{mahler2014advances} but, in this paper, a distance does not necessarily meet the metric properties.} should be a mathematically consistent metric on the selected space and, therefore, it meets the properties of non-negativity, identity, symmetry and triangle inequality \cite{Apostol_book74}\cite[Sec. 6.2.1]{mahler2014advances}. 

Besides the above fundamental properties, there are MTT-specific features that should be quantified in the metric on the space of sets of trajectories. For the closely related problem of multi-target filtering, which aims to estimate the current set of targets without forming trajectories, the optimal sub-pattern assignment (OSPA) metric \cite{Schuhmacher08_b,schuhmacher2008consistent} has played an important role over the past years. Given two sets of targets, OSPA matches all the targets in the smallest set to different targets in the other set to define a localization error. The rest of the targets in the largest set are penalized as cardinality error. However, traditional MTT performance assessment has been based on different concepts such as localization error for properly detected targets and costs for missed targets and false targets \cite[Sec. 13.6]{blackman1999design}, \cite{fridling1991performance,rothrock2000performance,mabbs1993performance,Drummond92}. These aspects have been quantified in a mathematically consistent way by the generalized OSPA (GOSPA) metric \cite{rahmathullah2016Generalized}. The GOSPA metric also avoids the spooky effect that appears in optimal estimation of multiple targets using the OSPA metric \cite{Angel19_d}.

For metrics on sets of trajectories, besides the above mentioned costs at each time step, there is the additional challenge posed by the temporal dimension of the trajectories. For instance, it is possible that, for a single trajectory in the ground truth, we get multiple estimated trajectories that form the best assignment at different time steps, due to their closer distance. This is referred to as track switching in the literature and should be penalised \cite[Sec. 13.6]{blackman1999design}. 

In the following, we proceed to review several distance functions to evaluate MTT algorithms \cite{vu2014new, lau2013tracking,schuhmacher2008consistent, ristic2011metric,bernardin2008evaluating, ganti1999clustering, kleinberg2002approximation, yianilos1993data, bento2016metric}. Though these distance functions are not defined on the space of sets of trajectories \cite{garcia2015setsTraj}, it is straightforward to extend the ideas to sets of trajectories, and we discuss these distances in the context of the space of sets of trajectories for comparison. 

The OSPA for tracks (OSPA-T) \cite{ristic2011metric} distance function was proposed as an extension of OSPA to sequences of sets of labeled targets, whose states include unique labels besides the physical state. However, OSPA-T returns counter-intuitive results \cite{vu2014new, lau2013tracking} and is not a metric \cite{vu2014new}. Another distance function that handles track switches is OSPA with track swaps (OSPA-TS), proposed in \cite[Sec. IV]{lau2013tracking}, but it is limited to a fixed and known number of trajectories with equal lengths. Another related distance function is the OSPA for multiple tracks (OSPA-MT) \cite{vu2014new}, but this function does not have a clear interpretation in terms of track switches, localization error, and missed and false targets. In computer vision, distance functions that penalize track switches are commonly used \cite{bernardin2008evaluating,ganti1999clustering, kleinberg2002approximation, yianilos1993data}. The most popular distance function in this field is called the classification of events, activities and relationships for multi-object tracking (CLEAR MOT) \cite{bernardin2008evaluating}. CLEAR MOT is not a metric and penalises track switches in a heuristic manner. 

As an improvement of the aforementioned cost functions, Bento and Zhu made several contributions in \cite{bento2016metric}. They first propose a family of distances, which are very general and are not required to be metrics. The switching cost in this family has the advantage that is defined very generally. We refer to this family as Bento's family of distances. Then, Bento and Zhu define a 'natural and computable' distance, which is computable in polynomial time using linear programming (LP), and then provide conditions for these distances to be metrics \cite[Thm. 5]{bento2016metric}. 
%However, we demonstrate in Appendix \ref{sec:Append_Bento_metric} that additional conditions are required to guarantee that the distances are metrics. 
We refer to the metrics that meet the conditions described in \cite[Thm. 5]{bento2016metric} as Bento's family of metrics, or simply Bento's metrics. Note that Bento's family of distances contains metrics that do not belong to Bento's family of metrics. 

In all Bento's metrics, $\ast$-trajectories are added to both the ground truth and the estimated set so that both sets have the same cardinality. Then, all trajectories in one set are assigned to trajectories in the other set via permutations and the switches are based on how these permutations change over time. Bento's metrics are flexible in the choice of switching penalty, which can be chosen depending on the application at hand. Nevertheless, calculating switches based on permutations of both real and $\ast$-trajectories is inherent in Bento's metrics and may provide counter-intuitive results, if the aim is to count the number of track switches, as illustrated in Section~\ref{subsec:motiv}. Another limitation of Bento's metrics is the lack of the parameter $p$ in the OSPA/GOSPA metrics, which is important to design optimal estimators  \cite{guerriero2010shooting, baum2015wasserstein,williams2015efficient}.  

In this paper, we propose a mathematically consistent metric on the space of finite sets of trajectories, in which the objective is to penalize localization errors for properly detected targets, costs for false and missed target, as well as the number of track switches \cite{Colegrove96, Colegrove06}.  We do so by extending the GOSPA metric (with parameter $\alpha=2$) \cite{rahmathullah2016Generalized} to trajectories where targets in one set are assigned to targets in the other set if they are properly detected, but they are left unassigned if they correspond to a missed or a false targets. Track switches are based on the changes in the assignments/unassignments of trajectories across time and are penalised based on their number, which requires the introduction of the concept of half-switches. It should be noted that Bento's metrics use the associations to $\ast$-trajectories to represent unassigned targets, but the switching cost is based on the change in permutations across time, not the change of assignments/unassignments across time. The two approaches are not equivalent as the mapping from permutations to assignments/unassignments also depends on the considered sets of trajectories. We also show in Appendix \ref{sec:Append_Bento_metric} that the proposed metric does not belong to Bento's family of metrics, though this family may be generalised to include our metric, e.g., by treating switches between different indices differently, as suggested in \cite{bento2016metric}.

To compute the proposed metric, we need to solve a multi-dimensional assignment problem \cite{garey2002computers, burkard2009assignment}, which can be efficiently solved by the Viterbi algorithm \cite{forney1973viterbi} for problems with few trajectories. Inspired by Bento's LP metric, we propose a lower bound on our metric by relaxing the binary constraint in the metric definition, making the lower bound an LP. This lower bound is also a metric and, due to the LP property, it can be computed in polynomial time \cite{khachiyan1980polynomial}, which makes it applicable to sets with a large number of trajectories. Similar LP relaxations of more complex multi-dimensional assignment problems have also been used in multiple hypothesis tracking  \cite{Storms00,Coraluppi00, Fortunato07}. 

A characteristic of all the above-mentioned metrics is that they assume that the ground truth and the estimate are known sets of trajectories. However, when we evaluate algorithms using the Bayesian framework via Monte Carlo simulations, these quantities are modelled as random variables. The final contribution of this paper is to extend the proposed metric to random finite sets (RFSs) of trajectories \cite{garcia2015setsTraj}, as was done in GOSPA for RFSs of targets  \cite{rahmathullah2016Generalized}. This is important for sound evaluation of algorithms using Monte Carlo simulation and to design optimal estimators. 

The outline of the paper is as follows. In Section~\ref{sec:prob}, we formulate the problem and discuss the challenges in designing a metric for sets of trajectories. Section~\ref{sec:assign} presents the proposed metric based on multi-dimensional assignments and, in Section~\ref{sec:lpReMetric}, we present the LP metric and its decomposition in terms of localization costs for properly detected targets, and costs for missed targets, false targets and track switches. We extend the metric to RFS of trajectories in Section~\ref{sec:extenRandomSets} and, in Section~\ref{sec:imple}, we analyze the proposed metric implementations via simulations. Finally, conclusions are drawn in Section~\ref{sec:concl}.

%%%%%%%%%%%%%%%%%% Example 1 %%%%%%%%%%%%%%%%%%%%%%%%%
\tikzset{x=0.8cm,y=0.8cm, every text node part/.style={align=center}, every node/.style={font=\scriptsize, inner sep=1pt,outer sep=0pt, minimum size=2pt,  draw}}
\begin{figure*}[th!]
\begin{center}
\begin{minipage}[t]{0.24\textwidth}
\centering
\begin{tikzpicture}
	\def \len {5}; 	\def \lenn {5};\def \del{0.5}; \def \ylim {-0.5};
	% X1
	\node [rectangle, label = below left:$X_{1}$](x1) at (1, 0){};
	\foreach \x in {2,..., \len} {
		\node [rectangle](x\x) at (\x, 0){};}
	\foreach \x in {2,..., \len} {
		\pgfmathtruncatemacro{\cur}{\x}
		\pgfmathtruncatemacro{\next}{\x - 1}
	\draw (x\cur)--(x\next);}
	
	%Y1
	\node [circle, label = above left:$Y_{1}$](y11) at (1, \del){};
	\foreach \x in {2,..., \lenn} {
			\pgfmathtruncatemacro{\te}{\x}
		\node [circle](y1\x) at (\x, \del){};}
	\foreach \x in {2,..., \lenn} {
		\pgfmathtruncatemacro{\cur}{\x}
		\pgfmathtruncatemacro{\next}{\x - 1}
	\draw [densely dotted](y1\cur)--(y1\next);}
	\draw [<->]([xshift=-10pt]x1.north west) -- ([xshift=-10pt]y11.south west) node[draw=none,fill=none,midway,left] {$\Delta$};
	  \foreach \x in {1, ..., \len}{
    \draw (\x , \ylim) -- (\x, \ylim-0.1) node[draw = none, below] {$\x$};}            
  \draw [-](1, \ylim) -- (\len, \ylim) node[draw=none, right] {$k$};
\end{tikzpicture}
%\subcaption{Assignment for $X_{1}$ is $(1, 1, 1, 1, 1)$.}
\subcaption{}
\label{fig:eg1a}
\end{minipage}
\begin{minipage}[t]{0.24\textwidth}
\centering
\begin{tikzpicture}
	\def \len {5}; 	\def \lenn {4};\def \del{0.5};\def \ylim {-0.5};
	% X1
	\node [rectangle, label = below left:$X_{1}$](x1) at (1, 0){};
	\foreach \x in {2,..., \len} {
		\node [rectangle](x\x) at (\x, 0){};}
	\foreach \x in {2,..., \len} {
		\pgfmathtruncatemacro{\cur}{\x}
		\pgfmathtruncatemacro{\next}{\x - 1}
	\draw (x\cur)--(x\next);}
	
	% Y1
	\node [circle, label = above left:$Y_{1}$](y11) at (1, \del){};
	\foreach \x in {1,..., \lenn} {
		\node [circle](y1\x) at (\x, \del){};}
	\foreach \x in {2,..., \lenn} {
		\pgfmathtruncatemacro{\cur}{\x}
		\pgfmathtruncatemacro{\next}{\x - 1}
	\draw [densely dotted](y1\cur)--(y1\next);}
	\draw [<->]([xshift=-10pt]x1.north west) -- ([xshift=-10pt]y11.south west) node[draw=none,fill=none,midway,left] {$\Delta$};
	  \foreach \x in {1, ..., \len}{
    \draw (\x , \ylim) -- (\x, \ylim-0.1) node[draw = none, below] {$\x$};}            
  \draw [-](1, \ylim) -- (\len, \ylim) node[draw=none, right] {$k$};
\end{tikzpicture}
%\subcaption{Assignment for $X_{1}$ is $(1, 1, 1, 1, 1)$.}
\subcaption{}
\label{fig:eg1b}
\end{minipage}
\begin{minipage}[t]{0.24\textwidth}
\centering
\begin{tikzpicture}
	\def \len {5}; 	\def \lenn {3};\def \del{0.5};\def \ylim {-0.5};
	% X1
	\node [rectangle, label = below left:$X_{1}$](x1) at (1, 0){};
	\foreach \x in {2,..., \len} {
		\node [rectangle](x\x) at (\x, 0){};}
	\foreach \x in {2,..., \len} {
		\pgfmathtruncatemacro{\cur}{\x}
		\pgfmathtruncatemacro{\next}{\x - 1}
	\draw (x\cur)--(x\next);}
	
	% Y1
	\node [circle, label = above left:$Y_{1}$](y11) at (1, \del){};
	\foreach \x in {1,..., \lenn} {
		\node [circle](y1\x) at (\x, \del){};}
	\foreach \x in {2,..., \lenn} {
		\pgfmathtruncatemacro{\cur}{\x}
		\pgfmathtruncatemacro{\next}{\x - 1}
	\draw [densely dotted](y1\cur)--(y1\next);}
	
	% Y2		
	\pgfmathtruncatemacro{\tmp}{\lenn+1};
	\node [circle, label = above left:$Y_{2}$](y2\tmp) at (\tmp, \del){};
	\pgfmathtruncatemacro{\tmp}{\lenn+2};
	\foreach \x in {\tmp, ..., \len} {
		\node [circle](y2\x) at (\x, \del){};}
	\foreach \x in {\tmp, ..., \len} {
		\pgfmathtruncatemacro{\cur}{\x}
		\pgfmathtruncatemacro{\next}{\x - 1}
	\draw [densely dotted](y2\cur)--(y2\next);}
	\draw [<->]([xshift=-10pt]x1.north west) -- ([xshift=-10pt]y11.south west) node[draw=none,fill=none,midway,left] {$\Delta$};
	  \foreach \x in {1, ..., \len}{
    \draw (\x , \ylim) -- (\x, \ylim-0.1) node[draw = none, below] {$\x$};}            
  \draw [-](1, \ylim) -- (\len, \ylim) node[draw=none, right] {$k$};
\end{tikzpicture}
%\subcaption{Assignment for $X_{1}$ is $(1, 1, 1, 2, 2)$.}
\subcaption{}
\label{fig:eg1c}
\end{minipage}
\begin{minipage}[t]{0.24\linewidth}
\centering
\begin{tikzpicture}
	\def \len {5}; 	\def \del{0.5}; \def \DEL{2};
	\def \swpt {4};\def \ylim {-0.5};
	\pgfmathtruncatemacro{\swptn}{\swpt-1}
	% X2
	\node [rectangle, label = above left:$X_{2}$](x21) at (1, \DEL){};
	\foreach \x in {2,..., \swptn} {
		\node [rectangle](x2\x) at (\x, \DEL){};}
	\foreach \x in {2,..., \swptn} {
		\pgfmathtruncatemacro{\cur}{\x}
		\pgfmathtruncatemacro{\next}{\x - 1}
	\draw (x2\cur)--(x2\next);}
	
	% X1
	\node [rectangle, label = below left:$X_{1}$](x11) at (1, 0){};
	\foreach \x in {2,..., \len} {
		\node [rectangle](x1\x) at (\x, 0){};}
	\foreach \x in {2,..., \len} {
		\pgfmathtruncatemacro{\cur}{\x}
		\pgfmathtruncatemacro{\next}{\x - 1}
	\draw (x1\cur)--(x1\next);}
	
	% Y3
	\node [circle, label = below right:$Y_{3}$](y31) at (1,  \DEL-\del){};
	\foreach \x in {2,..., \len} {
		\node [circle](y3\x) at (\x,  \DEL-\del){};}
	\foreach \x in {2,..., \len} {
		\pgfmathtruncatemacro{\cur}{\x}
		\pgfmathtruncatemacro{\next}{\x - 1}
		\draw [densely dotted](y3\cur)--(y3\next);}
	
	% Y1
	\node [circle, label = above right:$Y_{1}$](y11) at (1, \del){};
	\foreach \x in {2,..., \len} {
		\node [circle](y1\x) at (\x, \del){};}
%	\pgfmathtruncatemacro{\swptn}{\len-2}
	\foreach \x in {2,..., \swptn} {
		\pgfmathtruncatemacro{\cur}{\x}
		\pgfmathtruncatemacro{\next}{\x - 1}
		\draw [densely dotted](y1\cur)--(y1\next);}
	
	% Y2
	\pgfmathtruncatemacro{\swptp}{\swptn+1}
	\node [circle, label = above left:$Y_{2}$](y2\swptp) at (\swptp, \del){};	
	\pgfmathtruncatemacro{\swptpp}{\swptp+1}
	\foreach \x in {\swptpp,..., \len} {
		\node [circle](y2\x) at (\x, \del){};}
	\foreach \x in {\swptpp,..., \len} {
		\pgfmathtruncatemacro{\cur}{\x}
		\pgfmathtruncatemacro{\next}{\x - 1}
		\draw [densely dotted](y2\cur)--(y2\next);}
		
	% X3
	\pgfmathtruncatemacro{\swptp}{\swptn+1}
	\node [rectangle, label = above left:$X_{3}$](x3\swptp) at (\swptp,  \DEL){};	
	\pgfmathtruncatemacro{\swptpp}{\swptp+1}
	\foreach \x in {\swptpp,..., \len} {
		\node [rectangle](x3\x) at (\x,  \DEL){};}
	\foreach \x in {\swptpp,..., \len} {
		\pgfmathtruncatemacro{\cur}{\x}
		\pgfmathtruncatemacro{\next}{\x - 1}
		\draw (x3\cur)--(x3\next);}
		
	\draw [<->]([xshift=-10pt]x21.south west) -- ([xshift=-10pt]y31.north west) node[draw=none,fill=none,midway,left] {$\Delta$};
	\draw [<->]([xshift=-10pt]y11.north west) -- ([xshift=-10pt]y31.south west) node[draw=none,fill=none,midway,left] {$\delta$};
	\draw [<->]([xshift=-10pt]y11.south west) -- ([xshift=-10pt]x11.north west) node[draw=none,fill=none,midway,left] {$\Delta$};		    
	  \foreach \x in {1, ..., \len}{
    \draw (\x , \ylim) -- (\x, \ylim-0.1) node[draw = none, below] {$\x$};}
  \draw [-](1, \ylim) -- (\len, \ylim) node[draw=none, right] {$k$};
\end{tikzpicture}
%\subcaption{Assignments for $X_{1}$, $X_{2}$ and $X_{3}$ are $(1, 1, 1, 2, 2)$, $(3, 3, 3, 0, 0)$ and $(0,0, 0, 3, 3)$.}%}
\subcaption{}
\label{fig:eg1d}
\end{minipage}
\caption{Examples to illustrate the switching cost: (a) no switch, (b) no switch but cost for missed target at time $5$, (c) one switch and (d) two switches.}
\label{fig:eg1}
\end{center}
\vspace{-0.5cm}
\end{figure*}

\section{Problem formulation and background}\label{sec:prob}
In this section, we formulate the problem of designing a metric for sets of trajectories, review the GOSPA metric and explain the challenges to design a suitable metric.
\subsection{Space and fundamental properties}\label{sub:Space}
Our objective is to design a metric on the space of finite sets of trajectories that has an intuitive interpretation and is computable in polynomial time. Below we unfold the problem.

In MTT, the ground truth and its estimate are collections of trajectories, where each trajectory is a sequence of states representing the evolution of the target states over time where the start and end times of the individual trajectories can vary.  Both the ground truth and the estimates can be represented as sets of trajectories \cite{garcia2015setsTraj}. In the set of trajectories representation, each trajectory $X \in \bX = \{X_{1}, \ldots, X_{\nx}\}$ is of the form $(\omega, x^{1:\nu})$, where $\omega \in \Z$ is the initial time of the trajectory, $\nu \in \Z$ is its length and $x^{1:\nu} = (x^{1}, \ldots, x^{\nu})$ denotes a finite sequence that contains target states $x^{1}, \ldots, x^{\nu} \in \RN$ at $\nu$ consecutive time steps starting from $\omega$. 
Given a single trajectory $X = (\omega, x^{1:\nu})$, the set  $\Tk(X)$ is the state of the trajectory at time step $k$ \cite{garcia2015setsTraj}:
\begin{flalign}\label{eq:tau_trajectory}
\Tk(X) \triangleq \begin{cases}
\{x^{k+1-\omega}\}  & \omega \leq k \leq \omega + \nu - 1 \\ \emptyset & \text{otherwise}
\end{cases}.
\end{flalign} 
In order to design the metric, we consider trajectories in the time interval from time $1$ to $T$. We therefore consider trajectories such that $\left(\omega,\nu\right)$ belongs to the set $I_{(T)}=\left\{ \left(\omega,\nu\right):1\leq\omega\leq T\,\mathrm{and}\,1\leq\nu\leq T-\omega+1\right\}$, and let $\Upsilon$ be the set of all finite sets of such trajectories. In this paper, when we refer to a set of trajectories we refer to a set of trajectories up to time step $T$.

The above-mentioned trajectory representation is designed to fit the standard RFS multi-target tracking (MTT) models, which contain birth and death events, but no possibility to resurrect \cite{mahler2007statistical}. Though not required for standard MTT models, which are the main focus of this paper, the representation can be easily generalised to handle trajectories with gaps, by representing single trajectories as $((t_1,x_1^{1:i_1}),..., (t_n,x_n^{1:i_n}))$, where $t_j$ is the start time of the $j$-th segment of the trajectory, $i_j$ its duration, which meets $t_j+i_j < t_{j+1}$,  and $n$ is the number of trajectory segments. The metric we propose is also valid for such representations by defining \eqref{eq:tau_trajectory} for this case and, like Bento's metrics, it can then handle sets of trajectories with gaps.

A metric on the space of sets of trajectories is a function $d\left(\cdot,\cdot\right):\ensuremath{\Upsilon}\times\ensuremath{\Upsilon}\rightarrow\left[0,\infty\right)$ that  satisfies the non-negativity, symmetry, identity and the triangle inequality \cite[Sec. 2.15]{rudin1964principles}. We emphasize here that the triangle inequality property, despite its abstractness, has a major practical importance in algorithm assessment \cite[Sec. 6.2.1]{mahler2014advances}. Suppose, for instance, that there are two estimates $\bY$ and $\bZ$ for a ground truth $\bX$, and that the metric indicates that the estimate $\bZ$ is close to both the ground truth $\bX$ and the other estimate $\bY$.  Then, according to intuition, the estimate $\bY$ should also be close to the ground truth $\bX$. This property is ensured by the triangle inequality. 

\subsection{GOSPA metric} \label{sub:GOSPA}
In this section, we review the GOSPA metric between two sets of targets  $\mathbf{x}=\left\{ x_{1},...,x_{n_{\mathbf{x}}}\right\}$  and $\mathbf{y}=\left\{ y_{1},...,y_{n_{\mathbf{y}}}\right\}$, as it will be a foundation of the metric for sets of trajectories. For $\alpha=2$ \cite{rahmathullah2016Generalized}, the GOSPA metric can be written in terms of an assignment set $\theta$ between sets $\left\{ 1,..,n_{\mathbf{x}}\right\}$  and $\left\{ 1,...,n_{\mathbf{y}}\right\}$. That is, $\theta\subseteq\left\{ 1,..,n_{\mathbf{x}}\right\} \times\left\{ 1,..,n_{\mathbf{y}}\right\}$  such that $\left(i,j\right), \left(i,j'\right)\in\theta$ implies $j=j'$ and $\left(i,j\right), \left(i',j\right)\in\theta$ implies $i=i'$. Let $\Gamma_{\mathbf{x},\mathbf{y}}$ be the set of all possible assignment sets.

\begin{defn} 
	Given a metric $d_{b}\left(\cdot,\cdot\right)$ in $\mathbb{R}^{N}$, a scalar $c>0$, and a scalar $p$ with $1\leq p<\infty$, the GOSPA metric ($\alpha=2$) between sets $\mathbf{x}$ and $\mathbf{y}$ is \cite[Prop.1]{rahmathullah2016Generalized}
	\begin{align}
	&	d\left(\mathbf{x},\mathbf{y}\right)\nonumber\\
	&=\min_{\theta\in\Gamma_{\mathbf{x},\mathbf{y}}}\left(\sum_{\left(i,j\right)\in\theta}d_{b}^{p}\left(x_{i},y_{j}\right)+\frac{c^{p}}{2}\left(n_{\mathbf{x}}+n_{\mathbf{y}}-2\left|\theta\right|\right)\right)^{1/p}. \label{GOSPA} 
	\end{align} 
\end{defn}
In \eqref{GOSPA}, the first term represents the localization error for assigned targets (properly detected targets) to the power of $p$. The second term represents the cost of unassigned targets, which correspond to missed and false targets, to the power of $p$. 

In the metric for sets of trajectories, we will make use of GOSPA for sets $\mathbf{x}$ and $\mathbf{y}$ with at most one element. For $\mathbf{x}$ and $\mathbf{y}$, with $\left|\mathbf{x}\right|\leq1$ and $\left|\mathbf{y}\right|\leq1$, the GOSPA metric \eqref{GOSPA} can be written as
\begin{align}\label{eq:baseMetric}
d\left(\mathbf{x},\mathbf{y}\right)	\triangleq\begin{cases}
\min\left(c,d_{b}\left(x,y\right)\right) & \mathbf{x}=\left\{ x\right\} ,\mathbf{y}=\left\{ y\right\} \\
0 & \mathbf{x}=\mathbf{y}=\emptyset\\
\frac{c}{2^{1/p}} & \mathrm{otherwise.}
\end{cases}
\end{align}
In addition, the corresponding optimal assignment in \eqref{GOSPA} is 
\begin{align}
\theta^{\star}	=\begin{cases}
\left\{ \left(1,1\right)\right\}  & \left|\mathbf{x}\right|=\left|\mathbf{y}\right|=1,\,d\left(\mathbf{x},\mathbf{y}\right)<c\\
\emptyset & \mathrm{otherwise.}
\end{cases} \label{eq:GOSPA_opt_assignment}
\end{align}
Therefore, the targets in $\mathbf{x}$ and $\mathbf{y}$ are assigned if $\left|\mathbf{x}\right|=\left|\mathbf{y}\right|=1$ and $d\left(\mathbf{x},\mathbf{y}\right)<c$, and otherwise unassigned.

\subsection{Challenges}\label{sub:Challenges}
Besides the fundamental properties explained in Section \ref{sub:Space}, there are specific features to be considered in metrics for sets of trajectories. The properties that apply to metrics for sets of targets, such as penalising localization errors for properly detected targets and costs due to missed and false targets, are also relevant for metrics on sets of trajectories \cite{rahmathullah2016Generalized}. However, there are additional challenges posed by the temporal connection of the target states in trajectories, which should also be addressed. Below, we discuss these challenges in detail using examples. We use the notation $\bX $ for the ground truth set and $\bY$ for the estimated set.   

In the space of finite sets of targets, the concepts of localization error, missed targets and false targets are important \cite{Drummond92}, and can be quantified by the GOSPA metric, see Section \ref{sub:GOSPA}. These concepts can be extended to sets of trajectories by considering the target states of the trajectories at each time instant and summing the GOSPA costs across time. Let us analyse this use of GOSPA using the examples in Figures~\ref{fig:eg1a} and~\ref{fig:eg1b}, in which states are uni-dimensional and there are two different estimates of target states $\bY= \{Y_{1}\}$ for the same ground truth $\bX =\{X_{1}\}$. Assuming $\Delta \ll c$, it can be observed that from time $1$ to $4$, the states of the ground truth in both the examples have identical localization costs. However, at time step $5$, the target in Figure~\ref{fig:eg1a} has been properly detected as before, whereas in Figure~\ref{fig:eg1b} it has been missed. If $\Delta \geq c$, the interpretation is that the target in $\bX$ has been missed at all the time steps and $\bY$ has false targets at all time steps in Figure~\ref{fig:eg1a} and from time $1$ to $4$ in Figure~\ref{fig:eg1b}.

Even though the concepts of localization error for properly detected targets, missed and false targets are relevant to sets of trajectories, it is not sufficient to use the sum of GOSPA costs across time as a cost function. We also need to take the temporal dimension of the trajectories into account, which leads to the difficult problem of penalizing track switches \cite{garcia2014bayesian, blom2008tracking, crouse2011developing}. In traditional MTT performance evaluation methods, track switches are usually counted \cite{Colegrove96, Colegrove06}, so our objective is to design a metric with this property. Below, we provide two examples to illustrate this.

Consider the examples in Figures~\ref{fig:eg1a} and~\ref{fig:eg1c}. We argue that the estimate in Figure~\ref{fig:eg1a} is better than the one in Figure~\ref{fig:eg1c} as the latter has estimated the trajectory in two parts, and, as $\Delta\rightarrow0$, $\bY$ and $\bX$ become identical in Figure~\ref{fig:eg1a} but not in Figure~\ref{fig:eg1c}. However, the sum of GOSPA costs across time yields the same localization error for both estimates. The problem is that the localization cost does not consider how trajectories are connected across time, which prevents it from penalizing $\bY$ for splitting tracks in Figure~\ref{fig:eg1c}. 

Now, consider the examples in Figure~\ref{fig:eqcompnorm}. Assuming $\delta \gg c$, where $\delta$ is the distance indicated in the figure,  and $\Delta \ll c$, from a point of view of counting the number of track switches, we argue that the estimate $\bY$ in Figure~\ref{fig:egcompnorm_b} is better than the estimate $\bY$ in Figure~\ref{fig:egcompnorm_a}. The reason is that both estimates provide the same localization costs at all time steps. However, in Figure~\ref{fig:egcompnorm_a}, both $X_{1}$ and $X_{2}$ change association of estimates across time while, in Figure~\ref{fig:egcompnorm_b}, only $X_{2}$ changes association. Nevertheless, in some applications, it may be more important to penalise the track fragmentation in Figure~\ref{fig:egcompnorm_b} more than the double track switch in Figure~\ref{fig:egcompnorm_a}.

There are other desirable properties, besides the above, that the metric should have for the MTT applications we have in mind.  Other applications may have different desirable properties. First, if we flip the time axis, or translate both sets of trajectories in time or space, without changing the distance between them, the costs should remain the same. Second, it is useful that the metric satisfies a clustering property, which enables efficient metric computation in large problems, see Section \ref{sub:clustering}. That is, if we have a pair of sets $\mathbf{X}_{1}$ and $\mathbf{Y}_{1}$ in one area, and another pair of sets $\mathbf{X}_{2}$ and $\mathbf{Y}_{2}$ in a far away area, then, the metric (to the $p$-th power) between $\mathbf{X}=\mathbf{X}_{1}\cup\mathbf{X}_{2}$ and $\mathbf{Y}=\mathbf{Y}_{1}\cup\mathbf{Y}_{2}$ should be the sum of the metrics (to the $p$-th power) between $\mathbf{X}_{1}$ and $\mathbf{Y}_{1}$, and $\mathbf{X}_{2}$ and $\mathbf{Y}_{2}$. In particular, this implies that the costs (to the $p$-th power) for localization, false targets, missed targets and track switches  between $\mathbf{X}$ and $\mathbf{Y}$ should be the sum of the corresponding costs (to the $p$-th power) between $\mathbf{X}_{1}$ and $\mathbf{Y}_{1}$, and $\mathbf{X}_{2}$ and $\mathbf{Y}_{2}$. This is intuitively appealing since the overall estimation error then aggregates the error in well separated regions. The clustering property implies that assuming  $\delta \gg c$ and $\Delta \ll c$, the cost in Figure~\ref{fig:eg1d} should be double the cost (for $p=1$) in Figure~\ref{fig:eg1c}. The reason is that Figure~\ref{fig:eg1d} contains two clusters, and within each cluster, any metric provides identical values due to the symmetry property.

Finally, a metric for sets of trajectories must be independent of the indexing of the elements in $\bX$ and $\bY$. That is, these indices are not track IDs and, in all previous examples, we can interchange any $X_i$ with $X_j$, and $Y_i$ with $Y_j$, without affecting the metric.

\begin{figure}[h!]
\centering
\begin{minipage}{0.49\linewidth}
\centering
\begin{tikzpicture}
	\def \len {4}; 	\def \del{0.4}; \def \DEL{2}
	% X1
	\node [rectangle, label = below left:$X_{1}$](x11) at (1, 0){};
	\foreach \x in {2,..., \len} {
		\node [rectangle](x1\x) at (\x, 0){};}
	\foreach \x in {2,..., \len} {
		\pgfmathtruncatemacro{\cur}{\x}
		\pgfmathtruncatemacro{\next}{\x - 1}
	\draw (x1\cur)--(x1\next);}
	
	% X2
	\node [rectangle, label = above left:$X_{2}$](x21) at (1, \DEL){};
	\foreach \x in {2,..., \len} {
		\node [rectangle](x2\x) at (\x, \DEL){};}
	\foreach \x in {2,..., \len} {
		\pgfmathtruncatemacro{\cur}{\x}
		\pgfmathtruncatemacro{\next}{\x - 1}
	\draw (x2\cur)--(x2\next);}
	
	\def \swpt {3};
	% Y1
	\node [circle, label = above right:$Y_{1}$](y11) at (1, \del){};
	\foreach \x in {2,..., \len} {
		\node [circle](y1\x) at (\x, \del){};}
	
	% Y2
	\node [circle, label = below right:$Y_{2}$](y21) at (1, \DEL-\del){};
	\foreach \x in {2,..., \len} {
		\node [circle](y2\x) at (\x, \DEL - \del){};}
	
	% switches
	\pgfmathtruncatemacro{\swptn}{\swpt-1}
	\foreach \x in {2,..., \swptn} {
		\pgfmathtruncatemacro{\cur}{\x}
		\pgfmathtruncatemacro{\next}{\x - 1}
		\draw [densely dotted](y1\cur)--(y1\next);
		\draw [densely dotted](y2\cur)--(y2\next);}
		
	\pgfmathtruncatemacro{\cur}{\swpt}
	\pgfmathtruncatemacro{\next}{\swpt - 1}
	\draw [densely dotted](y1\cur)--(y2\next);	
	\draw [densely dotted](y2\cur)--(y1\next);
	
	\pgfmathtruncatemacro{\swptp}{\swpt+1}
	\foreach \x in {\swptp,..., \len} {
		\pgfmathtruncatemacro{\cur}{\x}
		\pgfmathtruncatemacro{\next}{\x - 1}
		\draw [densely dotted](y2\cur)--(y2\next);
		\draw [densely dotted](y1\cur)--(y1\next);}
		
	\draw [<->]([xshift=-10pt]x11.north west) -- ([xshift=-10pt]y11.south west) node[draw=none,fill=none,midway,left] {$\Delta$};
	\draw [<->]([xshift=-10pt]y21.south west) -- ([xshift=-10pt]y11.north west) node[draw=none,fill=none,midway,left] {$\delta$};
	\draw [<->]([xshift=-10pt]y21.north west) -- ([xshift=-10pt]x21.south west) node[draw=none,fill=none,midway,left] {$\Delta$};
	
\end{tikzpicture}
\subcaption{}%Assignments for $X_{1}$  and $X_{2}$ are $(1, 1, 2, 2)$ and $(2, 2, 1, 1)$ respectively.}
\label{fig:egcompnorm_a}
\end{minipage}
\begin{minipage}{0.49\linewidth}
\centering
\begin{tikzpicture}
	\def \len {4}; 	\def \del{0.4}; \def \DEL{2}
	% X1
	\node [rectangle, label = below left:$X_{1}$](x11) at (1, 0){};
	\foreach \x in {2,..., \len} {
		\node [rectangle](x1\x) at (\x, 0){};}
	\foreach \x in {2,..., \len} {
		\pgfmathtruncatemacro{\cur}{\x}
		\pgfmathtruncatemacro{\next}{\x - 1}
	\draw (x1\cur)--(x1\next);}
	
	% X2
	\node [rectangle, label = above left:$X_{2}$](x21) at (1, \DEL){};
	\foreach \x in {2,..., \len} {
		\node [rectangle](x2\x) at (\x, \DEL){};}
	\foreach \x in {2,..., \len} {
		\pgfmathtruncatemacro{\cur}{\x}
		\pgfmathtruncatemacro{\next}{\x - 1}
	\draw (x2\cur)--(x2\next);}
	
	\def \swpt {3};
	% Y1
	\node [circle, label = above right:$Y_{1}$](y11) at (1, \del){};
	\foreach \x in {2,..., \len} {
		\node [circle](y1\x) at (\x, \del){};}
	\foreach \x in {2,..., \len} {
		\pgfmathtruncatemacro{\cur}{\x}
		\pgfmathtruncatemacro{\next}{\x - 1}
		\draw [densely dotted](y1\cur)--(y1\next);}
	
	% Y2
	\node [circle, label = below right:$Y_{2}$](y21) at (1, \DEL-\del){};
	\foreach \x in {2,..., \len} {
		\node [circle](y2\x) at (\x, \DEL - \del){};}
	\pgfmathtruncatemacro{\swptn}{\len-2}
	\foreach \x in {2,..., \swptn} {
		\pgfmathtruncatemacro{\cur}{\x}
		\pgfmathtruncatemacro{\next}{\x - 1}
		\draw [densely dotted](y2\cur)--(y2\next);}

	% Y3
	\pgfmathtruncatemacro{\swptp}{\swptn+1}
	\node [circle, label = below left:$Y_{3}$](y3\swptp) at (\swptp, \DEL-\del){};	
	\pgfmathtruncatemacro{\swptpp}{\swptp+1}
	\foreach \x in {\swptpp,..., \len} {
		\node [circle](y3\x) at (\x, \DEL - \del){};}
	\foreach \x in {\swptpp,..., \len} {
		\pgfmathtruncatemacro{\cur}{\x}
		\pgfmathtruncatemacro{\next}{\x - 1}
		\draw [densely dotted](y3\cur)--(y3\next);}
		
	\draw [<->]([xshift=-10pt]x11.north west) -- ([xshift=-10pt]y11.south west) node[draw=none,fill=none,midway,left] {$\Delta$};
	\draw [<->]([xshift=-10pt]y21.south west) -- ([xshift=-10pt]y11.north west) node[draw=none,fill=none,midway,left] {$\delta$};
	\draw [<->]([xshift=-10pt]y21.north west) -- ([xshift=-10pt]x21.south west) node[draw=none,fill=none,midway,left] {$\Delta$};		
\end{tikzpicture}
\subcaption{}%}
\label{fig:egcompnorm_b}
\end{minipage}
\caption{Example to illustrate track switches, $\delta \gg c$ and $\Delta \ll c$. We argue that there are fewer track switches in (b) than in (a), since, in the latter, only $X_{2}$ in the ground truth changes association.} 
\label{fig:eqcompnorm}
\vspace{-0.5cm}
\end{figure}

\subsection{Bento's family of metrics}\label{subsec:motiv}

Among the distance functions for sets of trajectories that are available in the literature \cite{vu2014new, lau2013tracking,schuhmacher2008consistent, ristic2011metric, ganti1999clustering, yianilos1993data, kleinberg2002approximation, bento2016metric}, only Bento's metrics \cite{bento2016metric} address the problem of track switches for an unknown number of trajectories.

The main difference between Bento's metrics and the proposed metric is how they handle track switches so we proceed to explain the track switching penalty in \cite{bento2016metric}. Bento's metrics add $\ast$-trajectories to both the ground truth and the estimate so that both sets have the same cardinality. $\ast$-symbol states are also appended to the real trajectories at the time instants they do not exist, such that all trajectories with their $\ast$-extensions have the same length. Then, target states in the ground truth are associated with the targets in the estimate at every time step. This association is performed by permuting the indices of the trajectories in one set (including $\ast$-trajectories), and the switching cost only depends on how this permutation changes over time.  A joint optimisation over all possible permutations across all time steps gives the metric value. We proceed to discuss three relevant metrics in the family. 

We first consider a switching cost that returns the value one if there is a change in the permutations at two consecutive time steps and zero if there are no changes, which is referred to as $\mathcal{K}_{count}$ in \cite{bento2016metric}. We refer to the resulting metric as the B1 metric. The B1 metric is very important in \cite{bento2016metric} as it is the base for Bento's LP metrics, which are computable in polynomial time. Another alternative, which we refer to as the B2 metric and that also admits an LP version\footnote{Note that the LP version of B1 uses the matrix 1-norm and the LP version of B2 uses the component-wise matrix 1-norm, which must be divided by 2 to get the correspondence between switching penalties using permutation and matrix notation. The LP version of B2 is used in the simulations in \cite{bento2016metric}.}, is to sum the number of switches in the permutations at two consecutive time steps. This corresponds to the Hamming metric over permutations, where the Hamming metric between two permutations $\pi=\left(\pi_{1},...,\pi_{n}\right)$ and $\sigma=\left(\sigma_{1},...,\sigma_{n}\right)$ of $\left(1,...,n\right)$ is \cite{Diaconis_book88} 
\begin{align}
H\left(\pi,\sigma\right)	=\left|\left\{ i:\pi_{i}\neq\sigma_{i}\right\} \right|.
\end{align}
The use of the Cayley metric as the switching cost in Bento's metrics could be interesting in some applications but has the drawback that there is no corresponding LP relaxation so its use is limited to scenarios with a small number of trajectories.

A drawback of B1 and B2 is that they do not meet the clustering property, which is useful for fast computation in large problems. For B1, this follows directly from its definition. We proceed to explain why this is the case for B2. According to this property, we recall that the switching cost in Figure~\ref{fig:eg1d} should be twice the switching cost in Figure~\ref{fig:eg1c}. In Figure~\ref{fig:eg1c}, Bento's metrics add two $\ast$-trajectories, $X_{2}$ and $X_{3}$, to $\bX$ and, one, $Y_{3}$, to $\bY$. For B2, the optimal permutation from times 1 to 3 is [1,2,3] and from times 4 to 5 is [2,1,3], where the $i$th component of the permutation vector indicates the index of the element in $\bY$ associated with $X_{i}$ \cite{bento2016metric}\cite{schuhmacher2008consistent}. In Figure~\ref{fig:eg1d}, Bento's metrics add three $\ast$-trajectories to $\bX$ and three more to $\bY$. Then, for B2, the optimal permutation from times 1 to 3 is [1,3,2,4,5,6] and from times 4 to 5 is [2,1,3,4,5,6]. That is, in Figure ~\ref{fig:eg1c}, two elements of the optimal permutation change but, in Figure~\ref{fig:eg1d}, three elements of the optimal permutation change. In the B2 metric, Figure~\ref{fig:eg1d} has a switching cost that is 3/2 the switching cost of Figure~\ref{fig:eg1c}. Bento's metric with Cayley switching cost scales well in this example.

In addition, if instead of considering Figure~\ref{fig:eg1d}, we consider a scenario which includes two replicas of the trajectories in Figure~\ref{fig:eg1c} in distant regions, the clustering property implies that this scenario has double switching cost compared to the scenario in Figure~\ref{fig:eg1c}, which is the case for B2. There may be applications, in which it is desired to penalise this scenario and Figure~\ref{fig:eg1d} differently, as B2 does.

In the case represented in Figure~\ref{fig:egcompnorm_a}, Bento's metrics add two $\ast$-trajectories to $\bX$ and, two more to $\bY$. Assuming that the switching cost is small, the optimal permutations for B2 at times 1 and 2 correspond to [1,2,3,4] and, at times 3 and 4, to [2,1,3,4]. We can see that there is a change in the permutation, from [1,2,3,4] to [2,1,3,4], which is penalised by B2. Let us consider Figure~\ref{fig:egcompnorm_b}. Now, Bento's metrics add three $\ast$-trajectories to $\bX$ and two to $\bY$. The optimal permutations for B2 are [1,2,3,4,5] at times 1 and 2 and [1,3,2,4,5] at times 3 and 4. B2, and also B1, therefore penalise both situations in the same way. This is not desired from a point of view of counting track switches, since there is no track switch for $X_1$ in Figure~\ref{fig:egcompnorm_b} but there is track switch for both $X_1$ and $X_2$ in Figure~\ref{fig:egcompnorm_a}. Nevertheless, other Bento's distances can penalise these examples differently, and, for some applications, it may be desired to penalise these two cases equally, as in B2.

%It turns out that, despite the fact that the situation in Figure~\ref{fig:egcompnorm_b} is arguably better than in Figure~\ref{fig:egcompnorm_a} from the point of view of counting track switches, 

\section{Metric for sets of trajectories based on multi-dimensional assignments} \label{sec:assign}
In Section \ref{sub:multi-dimensional_metric}, we present a metric for sets of trajectories, based on the multi-dimensional assignment problem \cite{burkard2009assignment}, that penalises localization costs for properly detected targets, missed and false targets, and track switches. In Section \ref{subsec:over}, we explain how this metric addresses the examples in Section \ref{sub:Challenges}. Section \ref{subsec:MDcomp} discusses how to compute the metric.

\subsection{Multi-dimensional assignment metric}\label{sub:multi-dimensional_metric}
In this section, we present the metric for sets of trajectories based on multi-dimensional assignments, which is given in Definition~\ref{def:mdmetric}. We first introduce some additional notation.

We use $\mathbf{x}_{i}^{k}$ and $\mathbf{y}_{j}^{k}$ to denote the sets of targets that describe the state of the trajectories $X_{i}$ and $Y_{j}$ at time step $k$, respectively. That is,
\begin{align}
\mathbf{x}_{i}^{k}	&=\tau^{k}\left(X_{i}\right)\\
\mathbf{y}_{j}^{k}	&=\tau^{k}\left(Y_{j}\right)
\end{align}
where $\tau^{k}\left(\cdot\right)$ is defined in \eqref{eq:tau_trajectory}. Note that these sets contain at most one element.

In the proposed metric, each target set $\mathbf{x}_{i}^{k}$ can be assigned to another target set $\mathbf{y}_{j}^{k}$ at the same time step or be left unassigned. We use $\Pi_{\bX, \bY}$ to denote the set of all possible assignment vectors  between the  index sets $\{1, \ldots, \nx\}$ and $\{0, \ldots, \ny\}$. That is, the assignment vector $\pi^k= [\pi_1^k , ... , \pi_{n_X}^k]\in \Pi_{\bX, \bY}$ at time $k$ is a vector $\pi^k \in \{0, \ldots, \ny\}^{\nx}$ such that its $i^{\text{th}}$ component $\pi^k_{i} = \pi^k_{i^{\prime}} = j > 0$ implies that $i=i^{\prime}$. Here, $\pi^{k}_{i}=j \neq 0$ implies that trajectory $i$ in $\bX$ is assigned to trajectory $j$ in $\bY$ at time $k$ and $\pi^{k}_{i}=0$ implies that trajectory $i$ in $\bX$ is unassigned at time $k$. 

The above definition of assignment vectors ensures that no two distinct indexes in $\{1, \ldots, \nx\}$ are assigned to the same $j \in \{1, \ldots, \ny\}$. However, multiple indexes in $\{1, \ldots, \nx\}$ can be assigned to the index $0$ implying that the corresponding trajectories are unassigned. Let $\tilde{\pi}^k \subseteq \{1, \dots, \ny\}$ denote the set of indexes of $\bY$ that are left unassigned, according to $\pi^k$. The multi-dimensional assignment metric is then defined as follows.

\begin{defn}\label{def:mdmetric}
	For $1\leq p < \infty$, cut-off parameter $c>0$, switching penalty $\gamma > 0$ and a base metric $\dbase(\cdot,\cdot)$ in the single target space $\mathbb{R}^{N}$, the multi-dimensional assignment metric $\dS(\bX, \bY)$ between two sets $\bX$ and $\bY$ of trajectories is
	\begin{flalign}
	\dS(\bX, \bY)& \triangleq \min\limits_{\substack{\pi^{k}\in \Pi_{\bX, \bY}\\ k = 1,\ldots, T}}  \Bigg(\sum_{k=1}^{T} d^{k}_{\bX, \bY}(\bX, \bY, \pi^{k})^p\nn\\
	& \qquad\qquad   +  \sum_{k=1}^{T-1} s_{\bX, \bY}\big(\pi^{k}, \pi^{k+1}\big)^p \Bigg)^{\frac{1}{p}} \label{eq:mdmetric}
	\end{flalign}
	where the costs (to the $p$-th power) for properly detected targets, missed targets and false targets at time step $k$ are

\begin{align}
d_{\mathbf{X},\mathbf{Y}}^{k}\left(\mathbf{X},\mathbf{Y},\pi^{k}\right)^{p}	&=\sum_{\left(i,j\right)\in\theta^{k}\left(\pi^{k}\right)}d\left(\mathbf{x}_{i}^{k},\mathbf{y}_{j}^{k}\right)^{p}
\, \nonumber \\
&+\frac{c^{p}}{2}\left(\left|\tau^{k}\left(\mathbf{X}\right)\right|+\left|\tau^{k}\left(\mathbf{Y}\right)\right|-2\left|\theta^{k}\left(\pi^{k}\right)\right|\right) \label{eq:mdloc2}
\end{align}
with 
\begin{align}
\theta^{k}\left(\pi^{k}\right)	&=\left\{ (i,\pi_{i}^{k}):i\in\{1,\ldots,\nx\},\vphantom{d\left(\mathbf{x}_{i}^{k},\mathbf{y}_{\pi_{i}^{k}}^{k}\right)}\right.\nonumber \\
&\,\left.\left|\mathbf{x}_{i}^{k}\right|=\left|\mathbf{y}_{\pi_{i}^{k}}^{k}\right|=1,\,d\left(\mathbf{x}_{i}^{k},\mathbf{y}_{\pi_{i}^{k}}^{k}\right)<c\right\} \label{eq:theta_assignment}
\end{align}
and the switching cost (to the $p$-th power) from time step $k$ to $k+1$ is given by
	\begin{equation}
	s_{\bX, \bY}(\pi^{k}, \pi^{k+1})^p  = \gamma^{p}\sum_{i=1}^{\nx} s\big(\pi_{i}^{k}, \pi_{i}^{k+1}\big) \label{eq:mdswi}
	\end{equation}
	\begin{flalign}
	s\big(\pi_{i}^{k}, \pi_{i}^{k+1}\big) &=  \begin{cases}
	0 & \pi_{i}^{k} = \pi_{i}^{k+1} \\
	1 & \pi_{i}^{k} \neq  \pi_{i}^{k+1}, \pi_{i}^{k} \neq 0, \pi_{i}^{k+1} \neq 0\\
	\frac{1}{2}  & \text{otherwise.  \quad\quad\quad\quad\quad\quad\quad\quad}  
	\end{cases} 
	\label{eq:mdswi_explicit}
	\end{flalign} 	
\end{defn}
Equation \eqref{eq:mdloc2} is the GOSPA metric to the $p$-th power without the minimization, see \eqref{GOSPA}. Instead of minimizing over a target-level association $\theta^k$, the target-level association is determined by the trajectory-level association $\pi^k$, while excluding associations between pairs of targets whose distance is larger than $c$, see \eqref{eq:theta_assignment}. It should be noted that, for $\left(i,j\right)\in\theta^{k}$, $\mathbf{x}_{i}^{k}$ and $\mathbf{y}_{j}^{k}$ contain precisely one element and their distance is smaller than $c$, so $d\left(\mathbf{x}_{i}^{k},\mathbf{y}_{j}^{k}\right)$ coincides with $d_{b}\left(\cdot,\cdot\right)$ evaluated at the corresponding vectors, see \eqref{eq:baseMetric}, which corresponds to the localization error. Therefore, \eqref{eq:mdloc2} represents the sum of the costs (to the $p$-th power) that correspond to localization error for properly detected targets (indicated by the assignments in $\theta^{k}\left(\pi^{k}\right)$), number of missed targets $(\left|\tau^{k}\left(\mathbf{X}\right)\right|-\left|\theta^{k}\left(\pi^{k}\right)\right|)$ and false targets $(\left|\tau^{k}\left(\mathbf{Y}\right)\right|-\left|\theta^{k}\left(\pi^{k}\right)\right|)$ at time step $k$. 

The term \eqref{eq:mdswi} represents the switching cost (to the $p$-th power) between time steps $k$ and $k+1$. The term $ s\big(\pi_{i}^{k}, \pi_{i}^{k+1}\big) $ corresponds to no switch when there are no changes in the assignments, a full switch when there is change from one non-zero to another non-zero assignment, and a half switch when there is a change from a zero to a non-zero assignment or vice versa. The concept of half switches is an intrinsic part of a metric that penalises track switches based on assignments/unassignments to ensure the symmetry of the metric, as will be illustrated in Section~\ref{subsec:over}. The parameter $\gamma$ is the switching penalty. The larger the value of $\gamma$ is, the higher a track switch costs. 

Therefore, the metric $\dS(\cdot, \cdot)$ in \eqref{eq:mdmetric} consists of $p$-th root of the sum of the cost (to the $p$-th power) for localization errors for properly detected targets, missed targets, false targets and switching costs across all time steps $k=1,...,T$, and the assignments that determine these costs are obtained by solving a joint optimization problem over all time steps. The proof that  $\dS(\cdot, \cdot)$ is a metric in the space $\Upsilon$ of sets of trajectories is provided as a special case of the proof of the LP metric, which is given by Proposition \ref{def:LPreMetric} in Section~\ref{sec:lpReMetric}.

\subsection{Examples}\label{subsec:over}
We proceed to illustrate how the proposed metric works in the examples of Section \ref{sub:Challenges}.
Let us first consider Figure~\ref{fig:eg1a}. The optimal assignments at each time step for $X_{1}$ are $(\pi^1,...,\pi^5)=(1,1,1,1,1)$, as $X_{1}$ is always associated to $Y_{1}$. There is no change in the assignments so there is no track switch, as desired. In Figure~\ref{fig:eg1c}, the assignments for $X_{1}$ are (1,1,1,2,2), as $X_{1}$ is associated to $Y_{1}$ during the first three time steps and to $Y_{2}$ later on. Clearly, there is a switch from index 1 to 2, which should be penalised.

Due to the symmetry property of metrics, the switching penalty must be the same if we consider the assignments from $\bY$ to $\bX$ or from $\bX$ to $\bY$. Interestingly, if we consider the assignments  $\bY$ to $\bX$ in Figure~\ref{fig:eg1c}, we have (1,1,1,0,0) for $Y_{1}$ and (0,0,0,1,1) for $Y_{2}$. Now, we have two switches from assigned to unassigned (or the other way round). As the value of the metric must be the same if we penalise the assignments of $\bX$ or $\bY$,  the cost of a switch that considers an unassignment must be half the cost of a switch from assigned to assigned and it is therefore referred to as half-switch. Equation \eqref{eq:mdswi_explicit} ensures that the proposed metric satisfies this property. Note that the fragmented estimated trajectory in Figure~\ref{fig:eg1c} is considered a track switch according to our metric definition.

In the proposed metric, the assignments for $X_{1}$ in Figure~\ref{fig:eg1b} are (1,1,1,1,1). It should be noted that, as in Bento's family, the proposed metric can also assign trajectories that no longer exist. We argue that these assignments capture the cases that an estimated trajectory has a delay or starts before the ground truth and is therefore not considered a track switch. In this case, our metric penalises localization error for the first four time steps, plus a missed target at time 5, but no track switch.

In Figure~\ref{fig:eg1d}, the assignments for $X_{1}$, $X_{2}$ and $X_{3}$ are (1,1,1,2,2), (3,3,3,0,0) and (0,0,0,3,3). We now have one switch and two half-switches, which make a total of two switches, which is double the switching cost in Figure~\ref{fig:eg1c}, as required for clustering. 

In  Figure~\ref{fig:egcompnorm_a}, the assignments for $X_{1}$ and $X_{2}$ are (1,1,2,2) and (2,2,1,1). Thus, there are two switches, one for $X_{1}$ and another for  $X_{2}$. In Figure~\ref{fig:egcompnorm_b}, the assignments for $X_{1}$ and $X_{2}$ are (1,1,1,1) and (2,2,3,3) so there is only one switch for $X_{2}$. Consequently, the proposed metric indicates that the estimate in Figure~\ref{fig:egcompnorm_b} is more accurate than the one in Figure~\ref{fig:egcompnorm_a}, as desired from a point of view of counting switches.

\subsection{Computation}\label{subsec:MDcomp}
The metric proposed in \eqref{eq:mdmetric} is computed by solving a multi-dimensional assignment problem. This problem can be solved using the Viterbi algorithm \cite{viterbi1967error, forney1973viterbi}, but it is only efficient for small problems (roughly $\nx, \ny\leq 10$ in MATLAB).  The Viterbi solution scales linearly with the duration $T$, which means that it is tractable to compute \eqref{eq:mdmetric} also for long trajectories, as long as $\nx$ and $\ny$ are small.

One can also use methods such as the dual decomposition \cite{sontag2011introduction, Komodakis11} to compute the assignment metric sub-optimally. Nevertheless, in the next section, we show that it is possible to get an accurate lower bound on the metric using linear programming which can be computed in polynomial time and is also a metric.

\section{LP metric for sets of trajectories}\label{sec:lpReMetric}
In this section, we first show that the metric in \eqref{eq:mdmetric} can be reformulated as an integer linear programming problem \cite{papadimitriou1982combinatorial} in Section \ref{subsec:binLP}. In Section \ref{sub:LP_metric}, we explain that when the integer constraints are relaxed, the result is an LP problem which provides a lower bound of the metric, can be computed in polynomial time, and is also a metric. We also explain the metric decomposition and how to reduce the computational burden in Sections \ref{sec:Decomposition} and \ref{sub:clustering}, respectively.
\subsection{Integer linear programming formulation}\label{subsec:binLP}
In order to present the integer linear programming formulation of the metric $\dS(\bX, \bY)$, we introduce an equivalent representation of the assignments vectors $\pi^k\in\Pi_{\bX, \bY}$ in \eqref{eq:mdmetric} using binary weight matrices. Let $\W_{\bX, \bY}$ be the set of all binary matrices $W$ of dimension $(\nx+1) \times (\ny+1)$, representing associations between \bX~and \bY, such that $W^k$ satisfies the following properties:
\begin{flalign}
\sum_{i=1}^{\nx+1} W^k(i,j) &= 1, \ j = 1, \ldots, \ny  \label{eq:binConst1}\\
\sum_{j=1}^{\ny+1} W^k(i,j) &= 1, \ i = 1, \ldots, \nx  \label{eq:binConst2}\\
W^k(\nx+1, \ny+1) &= 0,  \label{eq:binConst3}\\
W^k(i,j) &\in \{0, 1\}, \forall \ i ,j, \label{eq:binConst}
\end{flalign}
where $W^k(i,j)$ represents the element in the row $i$ and column $j$ of matrix $W^k$. Then,  $W^k(i,j)=1$ if $\mathbf{x}_{i}^{k}$ is associated to $\mathbf{y}_{j}^{k}$. If $\mathbf{x}_{i}^{k}$ is unassigned, $W^k(i,\ny+1)=1$. If $\mathbf{y}_{j}^{k}$ is unassigned, $W^k(\nx+1,j)=1$. The first two properties ensure that no two target indexes in $\{1, \ldots, \nx\}$ are assigned to the same $j$ and vice versa. 

There is a bijection between the sets $\Pi_{\bX, \bY}$ and $\W_{\bX, \bY}$, such that for $\pi^k\in\Pi_{\bX, \bY}$, $W^k\in\W_{\bX, \bY}$, $i = 1,\ldots,\nx$ and $j=1,\ldots,\ny$:
\begin{flalign}
&\pi^k_{i} = j \neq 0 &\Leftrightarrow &W^k(i,j) = 1 \label{eq:piijWij}\\
&\pi^k_{i} = 0 &\Leftrightarrow &W^k(i,\ny+1) = 1  \label{eq:pii0Wid}\\
&\nexists i \in \{1, \ldots, \nx\}, \pi_{i} = j \neq 0 &\Leftrightarrow &W^k(\nx+1, j) = 1.\label{eq:pi0jWdj}
\end{flalign}
To illustrate the above bijection, let us consider Figure~\ref{fig:eg1c},  where the assignment sequence of $X_{1}$ is $\pi = (1, 1, 1, 2, 2)$. The corresponding weight matrices $W^{k} \in\{0,1\}^{2\times 3}$  for time $k = 1, \ldots, 5$  are $W^{k}(1, 1) = 1$ for $k=1, 2, 3$, $W^{k}(1, 2)=1$ for $k=4,5$ and $W^{k}(i,j) = 0$ everywhere else. For the example in Figure~\ref{fig:eg1d}, $W^{k} \in\{0,1\}^{4\times 4}$ is such that $W^{k}(1, 1) = W^{k}(2, 3) = W^{k}(3, 4) = 1$ for $k=1, 2, 3$, $W^{k}(1, 2) = W^{k}(2, 4) =  W^{k}(3, 3) = 1$ for $k =4, 5$ and $W^{k}(i,j)= 0$ everywhere else.

\begin{lem}\label{lem:mdBinLp}
The multi-dimensional assignment metric $\dS(\cdot, \cdot)$ in \eqref{eq:mdmetric} between two sets $\bX$ and $\bY$ of trajectories   can be written as
\begin{flalign}
&\dS(\bX, \bY) = \min_{\substack{W^{k} \in \W_{\bX, \bY} \\ k = 1, \ldots, T}}\Bigg(\sum_{k=1}^{T} \tr\big[\big(D_{\bX,\bY}^{k}\big)^{\dagger}W^{k}\big]  &\nn\\
&\quad + \frac{\gamma^{p}}{2}\sum_{k=1}^{T-1}\sum_{i=1}^{\nx} \sum_{j=1}^{\ny} |W^{k}(i,j) - W^{k+1}(i,j)|\Bigg)^{\frac{1}{p}},& \label{eq:binLP}
\end{flalign}
where $D_{\bX,\bY}^{k}$ is a $(\nx+1) \times (\ny+1)$ matrix whose $(i,j)$ element is
\begin{equation}
D_{\bX,\bY}^{k}(i,j)=d\left(\mathbf{x}_{i}^{k},\mathbf{y}_{j}^{k}\right)^{p}  \label{eq:DMatBase}
\end{equation}
where $\mathbf{x}_{n_{\mathbf{X}}+1}^{k}=\emptyset$ and $\mathbf{y}_{n_{\mathbf{Y}}+1}^{k}=\emptyset$, $\tr(\cdot)$ is the matrix trace operator and $(\cdot)^{\dagger}$ denotes the matrix transpose.
\end{lem}

The proof of the above lemma follows immediately from the bijection defined between the sets $\Pi_{\bX, \bY}$ and $\W_{\bX, \bY}$ in \eqref{eq:piijWij}, \eqref{eq:pii0Wid} and \eqref{eq:pi0jWdj}, and noticing that  \eqref{eq:mdmetric} and \eqref{eq:binLP} provide identical localization and switching costs.

\subsection{LP metric}\label{sub:LP_metric}
In this section, we relax the binary constraints of matrices $W^{k}$  for $k=1,...,T$ in Lemma~\ref{lem:mdBinLp} and show that the result is a metric that is computable in polynomial time using linear programming. 

Let $\hW_{\bX, \bY}$ be the set of all matrices $W^k$ of dimension $(\nx+1) \times (\ny+1)$ such that $W^k$ satisfies \eqref{eq:binConst1}, \eqref{eq:binConst2}, \eqref{eq:binConst3} and
\begin{flalign}
W^k(i,j) \geq 0,& \quad \forall i ,j. \label{eq:LPConst2}
\end{flalign}
The main difference to $\W_{\bX, \bY}$ is the relaxation of the constraint in \eqref{eq:binConst} and so $\W_{\bX, \bY} \subset \hW_{\bX, \bY}$. The relaxation of the binary constraints can be interpreted as making soft assignments of trajectories from one set to the other. Below, we define a new distance function, $\dSh(\bX, \bY)$ where the only difference to $\dS(\bX, \bY)$ in \eqref{eq:binLP} is that the optimization is over $W^{k}$ in $\hW_{\bX, \bY}$ instead of $\W_{\bX, \bY}$. Therefore it follows immediately that $\dSh(\bX, \bY)  \leq \dS(\bX, \bY) $. 

\begin{prop}\label{def:LPreMetric}
For $1\leq p < \infty$, $c > 0$ and $\gamma >0$, the LP relaxation of metric $\dS(\bX, \bY)$ between sets $\bX$ and $\bY$ of trajectories is also a metric $\dSh(\bX, \bY)$, which is given by
\begin{flalign}
&\dSh(\bX, \bY) = \min_{\substack{W^{k} \in \hW_{\bX, \bY} \\ k = 1, \ldots, T}}\Bigg(\sum_{k=1}^{T} \tr\big[\big(D_{\bX,\bY}^{k}\big)^{\dagger}W^{k}\big]  &\nn\\
&\quad + \frac{\gamma^{p}}{2}\sum_{k=1}^{T-1}\sum_{i=1}^{\nx} \sum_{j=1}^{\ny} |W^{k}(i,j) - W^{k+1}(i,j)|\Bigg)^{\frac{1}{p}},& \label{eq:LPreMetric}
\end{flalign}
where the element $(i,j)$ of matrix $D_{\bX,\bY}^{k}$, $D_{\bX,\bY}^{k}(i,j)$, is given by \eqref{eq:DMatBase} and $W^{k} \in \hW_{\bX, \bY}$ is given by \eqref{eq:binConst1}, \eqref{eq:binConst2}, \eqref{eq:binConst3} and \eqref{eq:LPConst2}. 
\end{prop}

The proof that  $\dSh(\cdot, \cdot) $  is a metric on the space of sets of trajectories is provided in Appendix~\ref{sec:Append_metric}. In this appendix, we also prove that the metric is computable in polynomial time using LP \cite{khachiyan1980polynomial}. When the solution to the LP metric is given by integral matrices, it returns the same value as the multi-dimensional assignment metric, and therefore, has the same types of penalties for localization error for properly detected targets and costs for missed, false targets and track switches. For the examples in Figure~\ref{fig:eg1} and Figure~\ref{fig:eqcompnorm}, both the multi-dimensional assignment metric in \eqref{eq:mdmetric} and the LP metric in \eqref{eq:LPreMetric} provide the same values. It should be noted that the assignment matrices in Proposition \ref{def:LPreMetric} are of dimensions $(\nx+1) \times (\ny+1)$ which constitutes an important reduction in dimensionality over matrices in Bento's LP metrics \cite{bento2016metric}, whose dimensions are $(\nx+\ny) \times (\nx+\ny)$, for problems with many trajectories in both sets.

We would also like to remark that if $\gamma=\infty$, the feasible sets of \eqref{eq:mdmetric} and \eqref{eq:LPreMetric} meet $W^1=...=W^T$, and both \eqref{eq:mdmetric} and \eqref{eq:LPreMetric} become the same 2-D assignment problem, which has an integral solution. This problem is fast to solve compared to \eqref{eq:LPreMetric}, but does not allow for track switching, so it is not the most preferable choice to evaluate MTT algorithms.

\subsection{Metric decomposition}\label{sec:Decomposition}

In this section, we explain how the multi-dimensional assignment metric, with the formulation in Lemma \ref{lem:mdBinLp}, and the LP metric decompose into costs for properly detected targets, missed and false targets, and track switches. We first explain the decomposition for the multi-dimensional assignment metric. 

As explained after Definition \ref{def:mdmetric}, properly
detected targets in $\mathbf{X}$ are those assigned to an estimate in $\bY$ according to $\theta^k$ in \eqref{eq:mdloc2}. False targets are those
targets in $\mathbf{X}$ that are not assigned to an estimate in $\bY$ according to  $\theta^k$. Missed targets are those targets in $\mathbf{Y}$
that are not assigned to a target in $\bX$ according to $\theta^k$. Then, $d\left(\mathbf{x}_{i}^{k},\mathbf{y}_{j}^{k}\right)$ represents the following costs:
\begin{itemize}
	\item A localization error for a properly detected target if $i\leq n_{x},\,j\leq n_{y},\mathbf{x}_{i}^{k}\neq\emptyset,\mathbf{y}_{j}^{k}\neq\emptyset,d\left(\mathbf{x}_{i}^{k},\mathbf{y}_{j}^{k}\right)<c$. 
	\item A missed target if $i\leq n_{x},\,j\leq n_{y},\mathbf{x}_{i}^{k}\neq\emptyset,\mathbf{y}_{j}^{k}=\emptyset$
	or $i\leq n_{x},\,j=n_{y}+1$. 
	\item A false target if $i\leq n_{x},\,j\leq n_{y},\mathbf{x}_{i}^{k}=\emptyset,\mathbf{y}_{j}^{k}\neq\emptyset$
	or $i=n_{x}+1,\,j\leq n_{y}$.
	\item The sum of a missed and a false target cost, each with a cost $d\left(\mathbf{x}_{i}^{k},\mathbf{y}_{j}^{k}\right)^{p}/2$,
	if $i\leq n_{x},\,j\leq n_{y},\mathbf{x}_{i}^{k}\neq\emptyset,\mathbf{y}_{j}^{k}\neq\emptyset,d\left(\mathbf{x}_{i}^{k},\mathbf{y}_{j}^{k}\right)=c$. 
\end{itemize}
We denote the sets of indices $\left(i,j\right)$ that belong to each
of the previous categories at time step $k$ as $T_{1}^{k}$, $T_{2}^{k}$,
$T_{3}^{k}$ and $T_{4}^{k}$. Then, we have
\begin{align}
& d_{p}^{\left(c,\gamma\right)}(\mathbf{X},\mathbf{Y}) \nonumber\\
& =\min_{\substack{W^{k} \in \W_{\bX, \bY} \\ k = 1, \ldots, T}}\left(\sum_{k=1}^{T}\mathrm{l}^{k}\left(\mathbf{X},\mathbf{Y},W^{k}\right)^{p}\right.\nonumber\\
& \quad+\sum_{k=1}^{T}\mathrm{m}^{k}\left(\mathbf{X},\mathbf{Y},W^{k}\right)^{p}+\sum_{k=1}^{T}\mathrm{f}^{k}\left(\mathbf{X},\mathbf{Y},W^{k}\right)^{p}\nonumber\\
& \left.\quad+\sum_{k=1}^{T-1}\mathrm{s}^{k}\left(W^{k},W^{k+1}\right)^{p}\right)^{1/p} \label{eq:metric_decom}
\end{align}
where 
\begin{align*}
\mathrm{l}^{k}\left(\mathbf{X},\mathbf{Y},W^{k}\right)^{p} & =\sum_{\left(i,j\right)\in T_{l}^{k}}D_{\bX,\bY}^{k}(i,j)W^{k}\left(i,j\right)\\
\mathrm{m}^{k}\left(\mathbf{X},\mathbf{Y},W^{k}\right)^{p} & =\frac{c^{p}}{2}\sum_{\left(i,j\right)\in T_{2}^{k}\cup T_{4}^{k}}W^{k}\left(i,j\right)\\
\mathrm{f}^{k}\left(\mathbf{X},\mathbf{Y},W^{k}\right)^{p} & =\frac{c^{p}}{2}\sum_{\left(i,j\right)\in T_{3}^{k}\cup T_{4}^{k}}W^{k}\left(i,j\right)\\
\mathrm{s}^{k}\left(W^{k},W^{k+1}\right)^{p} & =\frac{\gamma^{p}}{2}\sum_{i=1}^{n_{\mathbf{X}}}\sum_{i=1}^{n_{\mathbf{Y}}}\left|W^{k}\left(i,j\right)-W^{k+1}\left(i,j\right)\right|
\end{align*}
represent the costs (to the $p$-th power)
for properly detected targets, missed targets, false targets and track
switches at time step $k$, given trajectory level assignments $W^{1:T}$. Note that the set $T_4^k$ appears in both $m^k(\cdot)$ and $f^k(\cdot)$, and one of its elements contributes $\frac{c^p}{2}$ to each, so its overall contribution is $c^p$. That is, the cost of a missed/false target is $\frac{c^p}{2}$. Once we compute the optimal assignment $W^{1:T}$ at a trajectory level,
we can report the decomposition of the metric in terms of these costs.
For the LP metric, we have the same decomposition but the assignments are soft.

\subsection{Computational aspects}\label{sub:clustering}

In this section, we explain two ways of accelerating metric calculation. We first proceed to explain clustering.

Trajectories $X_{i}$ and $Y_{j}$ will not be assigned to each other (at a trajectory and target level) at any time step in the final value of the metric if the following condition is met
\begin{itemize}
	\item $D_{\mathbf{X},\mathbf{Y}}^{k}(i,j)=c^{p}$ for all $k$ such that $\tau^{k}\left(X_{i}\right)\neq\emptyset$ and $\tau^{k}\left(Y_{j}\right)\neq\emptyset$.
\end{itemize}
The reason is that leaving these trajectories unassigned has the same cost as their assignment at a trajectory level (though in both cases trajectories are unassigned at a target level). Therefore, we can remove the possibility that these trajectories are assigned at a trajectory level as it does not affect the metric. Consequently, given $\mathbf{X}$ and $\mathbf{Y}$, we can compute the pairs of subsets in $\mathbf{X}$ and $\mathbf{Y}$ in which there can be assignments. One way to do this is to compute the adjacency matrix of the bipartite graph formed by the trajectories, using the condition stated above and determine the disjoint components, e.g., using the reverse Cuthill-McKee algorithm \cite{George_book81}. It should be noted that this type of spatial clustering cannot be applied to B1 and B2 metrics, see example in Figure \ref{fig:eg1d} discussed in Section \ref{subsec:motiv}.

In a given cluster, we can also speed up metric calculation by considering a shorter time window. In a cluster with sets of trajectories, $\mathbf{X}$ and $\mathbf{Y}$, we can calculate the minimum and maximum times at which there can be assignments (at a trajectory and target levels) using the criteria explained above. The parts of the trajectories that are outside this window provide a penalty of $c^{p}/2$ at the times at which they exist, without track switching penalty. That is, their assignments outside the considered time window are the ones at the window endpoints. Then, we only need to solve the optimisation problem in the time window in which there can be assignments. 
 
\section{Extension to random sets of trajectories} \label{sec:extenRandomSets}
In the previous sections, we studied metrics between deterministic finite sets of trajectories. However, in the Bayesian formulation of MTT, the ground truth is modelled as a random quantity and the estimates are sets that depend deterministically on the observed data \cite{mahler2007statistical}. In MTT performance evaluation using Monte Carlo simulation, the metric values are averaged over several realizations of the observed data and possibly the ground truth. In such scenarios, both the estimates and the ground truth can be interpreted as random finite sets, as they can change in each Monte Carlo run. It is therefore important to have a metric on the space of random sets of trajectories for performance evaluation. We proceed to  extend the proposed metrics for sets of trajectories to random finite sets of trajectories.

Before introducing the metric, we review the set integral for trajectories. Given a real-valued function $\pi\left(\cdot\right)$ on the single trajectory space, its integral is \cite{garcia2015setsTraj,Angel19_f}
\begin{equation}
\int\pi\left(X\right)dX	=\sum_{\left(\omega,\nu\right)\in I_{(T)}}\int\pi\left(\omega,x^{1:\nu}\right)dx^{1:\nu}.
\end{equation}
The single trajectory integral goes through all possible start times, lengths and target states of the trajectory. 

Given a real-valued function $\pi\left(\cdot\right)$ on the space of sets of trajectories, its set integral is \cite{garcia2015setsTraj}
\begin{equation}\label{set_integral_trajectory}
\int\pi\left(\mathbf{X}\right)\delta\mathbf{X}	=\sum_{n=0}^{\infty}\frac{1}{n!}\int\pi\left(\left\{ X_{1},...,X_{n}\right\} \right)dX_{1:n}
\end{equation} 
where $X_{1:n}=\left(X_{1},...,X_{n}\right)$. If $\pi\left(\cdot\right)$ is a multitrajectory density, then, $\pi\left(\cdot\right)\geq0$ and its set integral is one. An RFS of trajectories is uniquely characterised by its multitrajectory density. 

Let $\pi(\bX, \bY)$ be the joint multitrajectory density of the random sets of trajectories $\bX$ and $\bY$  \cite{mahler2014advances}. This joint multitrajectory density characterises the joint distribution of the two RFS $\bX$ and $\bY$. The expected value of the metric \eqref{eq:LPreMetric} to the power of $p$ is  
\begin{align}\label{Expected_error}
&	\mathbb{E}\left[\bar{d}_{p}^{\left(c,\gamma\right)}\left(\mathbf{X},\mathbf{Y}\right)^p\right] \nonumber\\
&=\int\int\bar{d}_{p}^{\left(c,\gamma\right)}\left(\mathbf{X},\mathbf{Y}\right)^p\pi\left(\mathbf{X},\mathbf{Y}\right)\delta\mathbf{X}\delta\mathbf{Y}\nonumber\\
&=\sum_{n=0}^{\infty}\sum_{m=0}^{\infty}\frac{1}{n!}\frac{1}{m!}\int\int\bar{d}_{p}^{\left(c,\gamma\right)}\left(\left\{ X_{1},...,X_{n}\right\} ,\left\{ Y_{1},...,Y_{m}\right\} \right)^p\nonumber\\
&\quad\times\pi\left(\left\{ X_{1},...,X_{n}\right\} ,\left\{ Y_{1},...,Y_{m}\right\} \right)dX_{1:n}dY_{1:m}.
\end{align}
Then, we prove in Appendix \ref{sec:Append_RFS_metric}, the following lemma.

\begin{lem}\label{lem:metricRFS}
Given $1\leq p'<\infty$, $\left(\E\big[\dSh(\bX, \bY)^{p'}\big]\right)^{1/p'}$ and $\left(\E\big[\dS(\bX, \bY)^{p'}\big]\right)^{1/p'}$ are metrics on the space of random finite sets of trajectories with finite moment $\mathbb{E}\left[\left|\cdot\right|^{p'/p}\right]$. 
\end{lem}

Lemma \ref{lem:metricRFS} considers that both RFS $\bX$ and $\bY$ of trajectories meet $\mathbb{E}\left[\left|\bX\right|^{p'/p}\right]< \infty$ and $\mathbb{E}\left[\left|\bY\right|^{p'/p}\right]< \infty$, which implies that the metric is finite. For $p'=p$, this condition implies that the RFSs have a finite mean number of trajectories.

In addition, for $p'=p$, we can decompose the metrics on RFS as follows. For each $\mathbf{X}$ and $\mathbf{Y}$, we denote $W_{\mathbf{X},\mathbf{Y}}^{k}$ the optimal assignment obtained when computing $d_{p}^{\left(c,\gamma\right)}(\mathbf{X},\mathbf{Y})$. Then, we can use the metric decomposition in \eqref{eq:metric_decom} to obtain
\begin{align}
&\ensuremath{\left(\mathbb{E}\big[d_{p}^{\left(c,\gamma\right)}(\mathbf{X},\mathbf{Y})^{p}\big]}\right)^{1/p} \nonumber\\&=\left[\sum_{k=1}^{T}\ensuremath{\left(\mathbb{E}\left[\mathrm{l}^{k}\left(\mathbf{X},\mathbf{Y},W_{\mathbf{X},\mathbf{Y}}^{k}\right)^{p}\right]+\mathbb{E}\left[\mathrm{m}^{k}\left(\mathbf{X},\mathbf{Y},W_{\mathbf{X},\mathbf{Y}}^{k}\right)^{p}\right]\right.}\right. \nonumber\\
&\left.+\ensuremath{\mathbb{E}\left[\mathrm{f}^{k}\left(\mathbf{X},\mathbf{Y},W_{\mathbf{X},\mathbf{Y}}^{k}\right)^{p}\right]}\right) \nonumber\\
&\left.+\sum_{k=1}^{T-1}\ensuremath{\mathbb{E}\left[\mathrm{s}^{k}\left(W_{\mathbf{X},\mathbf{Y}}^{k},W_{\mathbf{X},\mathbf{Y}}^{k+1}\right)^{p}\right]}\right]^{1/p}
\end{align}
where the first, second and third expectations are the average cost (to the $p$-th power) of the localization error for properly detected targets, missed targets and false targets at time step $k$, respectively, and the last expectation is the average switching cost from time step $k$ to $k+1$. 

The metrics on RFSs, and their decompositions, can be used to evaluate algorithms based on Monte Carlo runs, where the expectations are approximated by the average over the outputs. The decomposition is useful to analyse the differences between different filters, see for example \cite{garcia2015setsTraj,Angel19_f,Granstrom18,Xia19,Xia19_b}. The metrics on RFSs are also useful to compute optimal estimators \cite{rahmathullah2016Generalized,Angel19_d}. It should be noted that it usually aids to select $p'=p=2$ in Lemma \ref{lem:metricRFS} to obtain computable optimal estimators. In this case, when we set the Euclidean metric as the base metric $\dbase(\cdot, \cdot)$ on $\RN$ in \eqref{eq:baseMetric}, we get a sum of  squares form inside the  expectation. For the OSPA/GOSPA metrics with known target number, we can obtain the best estimator for $p'=p=2$ \cite{guerriero2010shooting}.

\section{Simulation results} \label{sec:imple}
In this section, we present simulation results to analyse the results of the metric for varying parameter values. The metric has also been used in \cite{garcia2015setsTraj,Granstrom18,Xia19,Xia19_b}, along with its decomposition, to assess and compare multiple target tracking algorithms\footnote{MATLAB code of the proposed LP metric is provided at https://github.com/Agarciafernandez/MTT.}.

\begin{figure}[ht!]
\centering
\begin{minipage}{0.49\linewidth}
	\centering
	\includegraphics[scale=0.25]{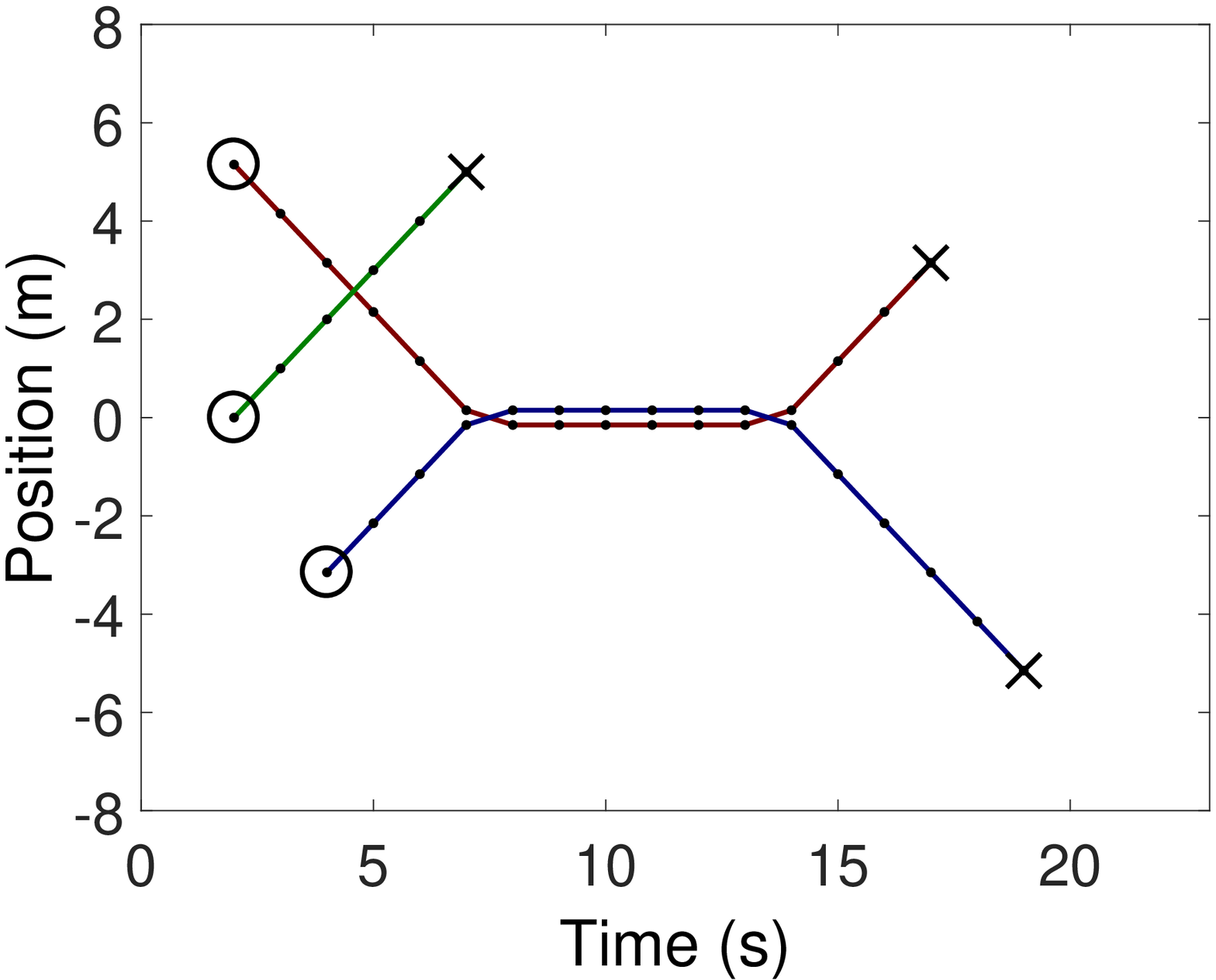}
	\subcaption{Ground truth $X$}
	\label{fig:case3Truth}
\end{minipage}
\begin{minipage}{0.49\linewidth}
	\centering
	\includegraphics[scale=0.24]{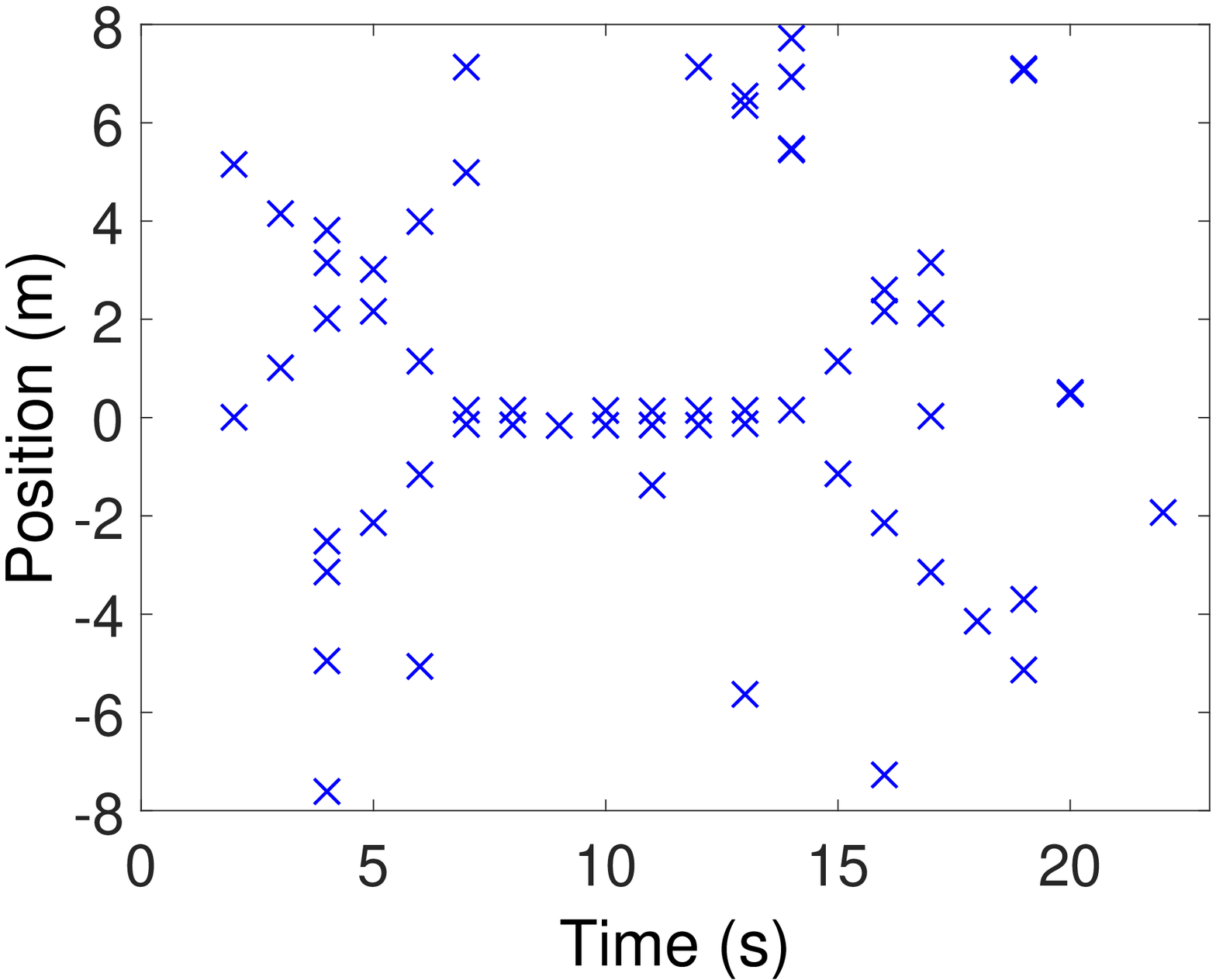}
	\subcaption{Measurements}
	\label{fig:meas}
\end{minipage}
\caption{The positional components of the ground truth (a) and the measurements (b) observed across time. The circles and crosses in (a) indicate the appearance and end times of the trajectories, respectively. }
\end{figure}

\begin{figure*}[!t]
	\begin{tabular}{c c c}
		\begin{minipage}{0.3\textwidth}
			\includegraphics[scale = 0.3]{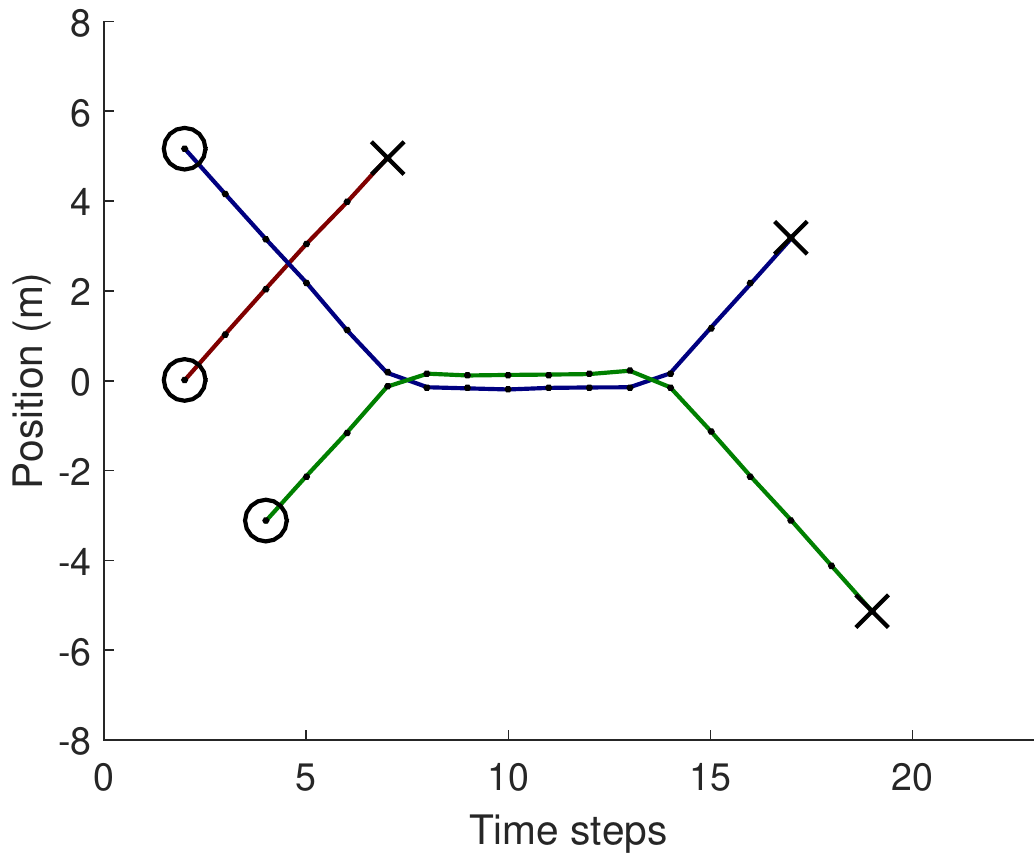}  \subcaption{$Y$ with localized states} \label{fig:case1}
		\end{minipage}& 
		\begin{minipage}{0.3\textwidth}
			\includegraphics[scale = 0.3]{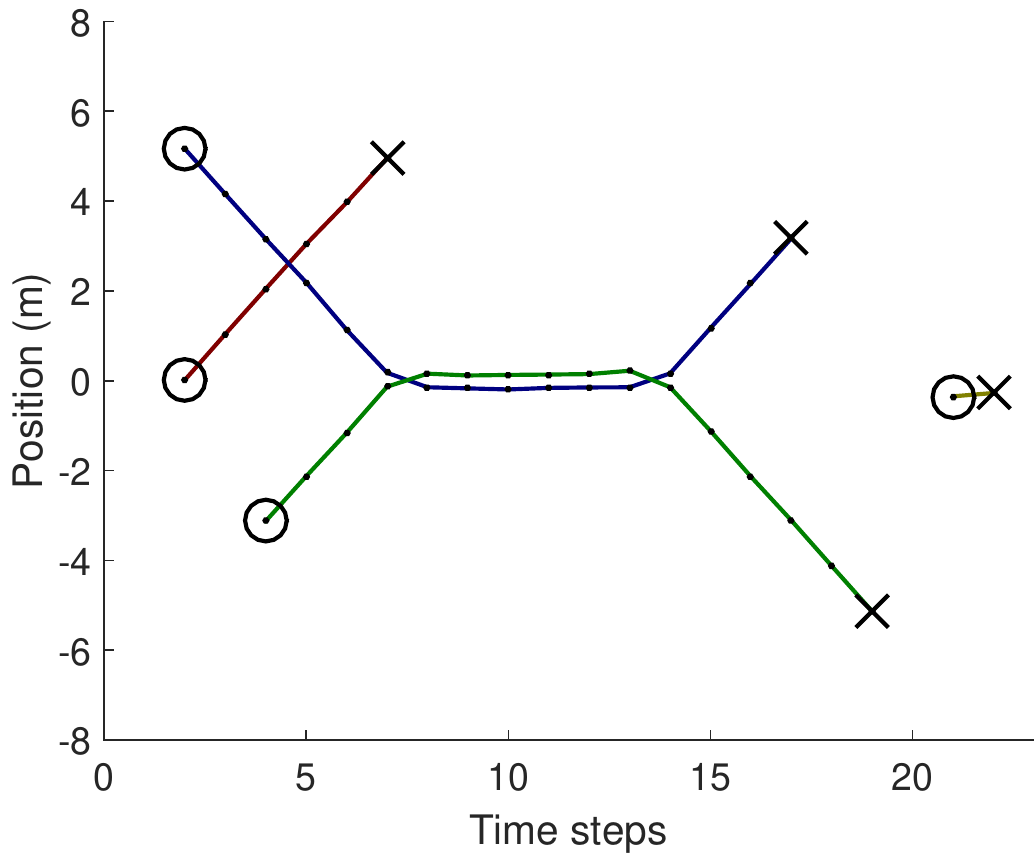}  \subcaption{$Y$ with localized states and a false track} \label{fig:case2}
		\end{minipage}& 
		\begin{minipage}{0.3\textwidth} 
			\includegraphics[scale = 0.3]{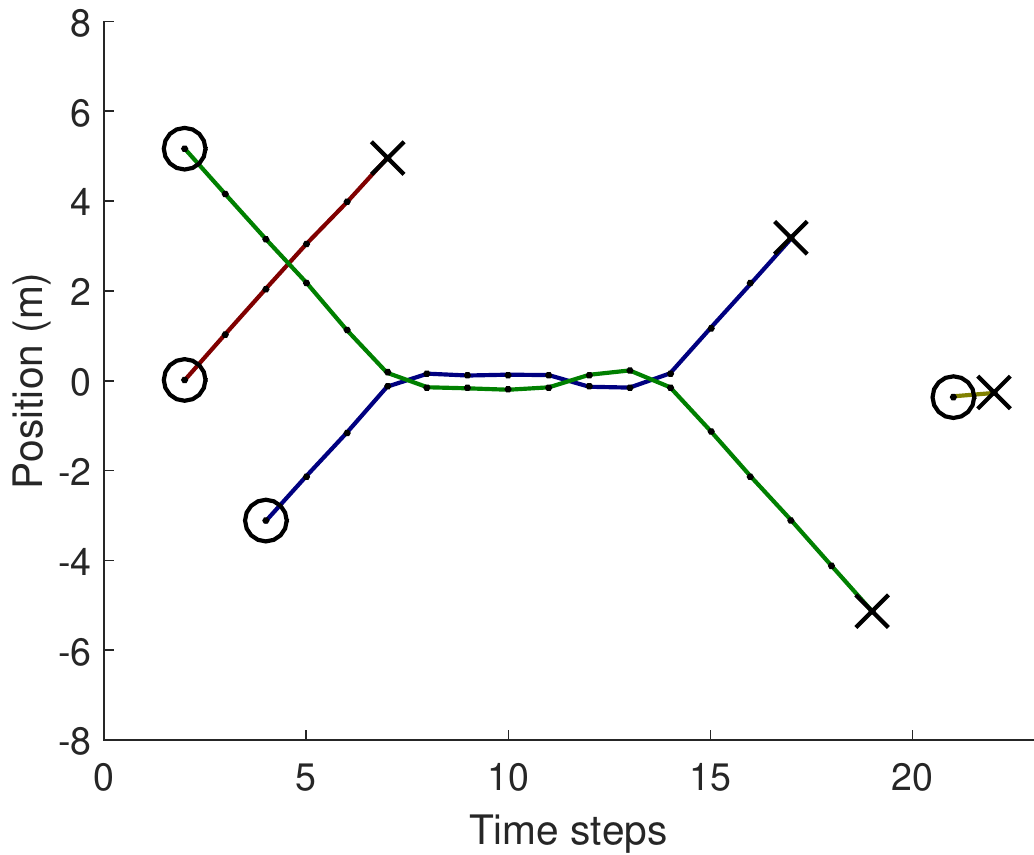} \subcaption{$Y$ with localized states, two switches and a false track} \label{fig:case3} 
		\end{minipage} \\
		\begin{minipage}{0.3\textwidth}  
			\includegraphics[scale = 0.3]{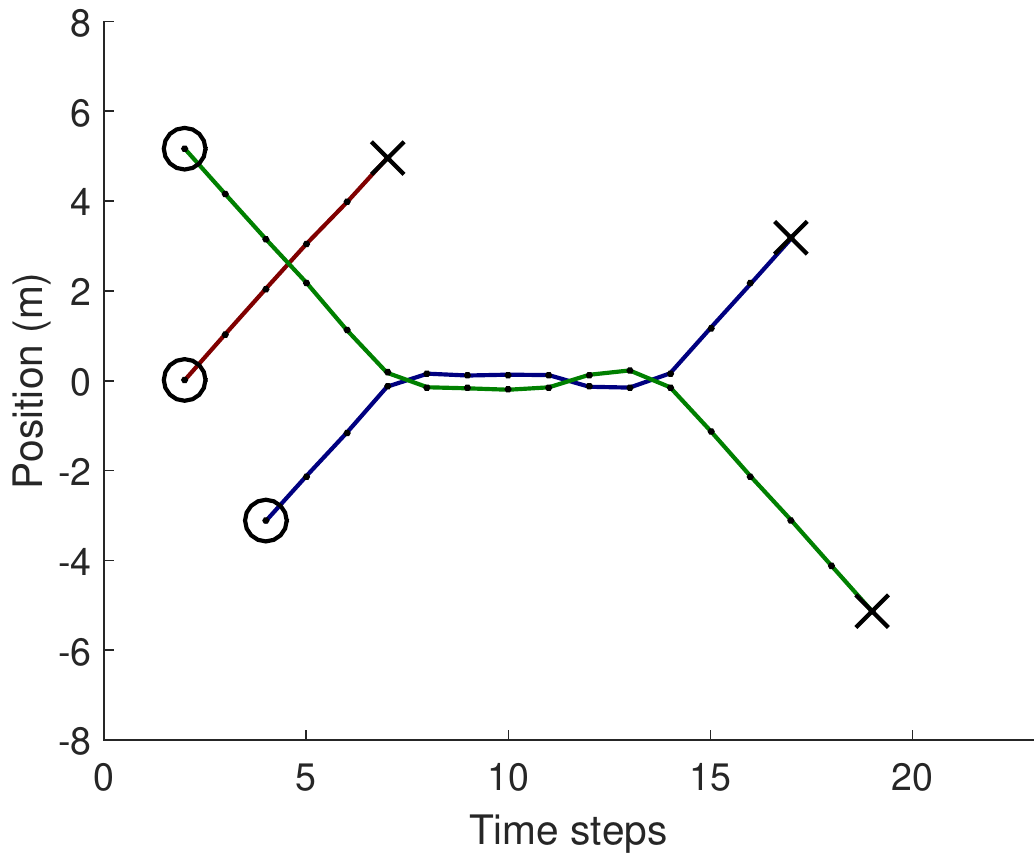} \subcaption{$Y$ with localized states and two  switches} \label{fig:case4} 
		\end{minipage}& 
		\begin{minipage}{0.3\textwidth} 
			\includegraphics[scale = 0.3]{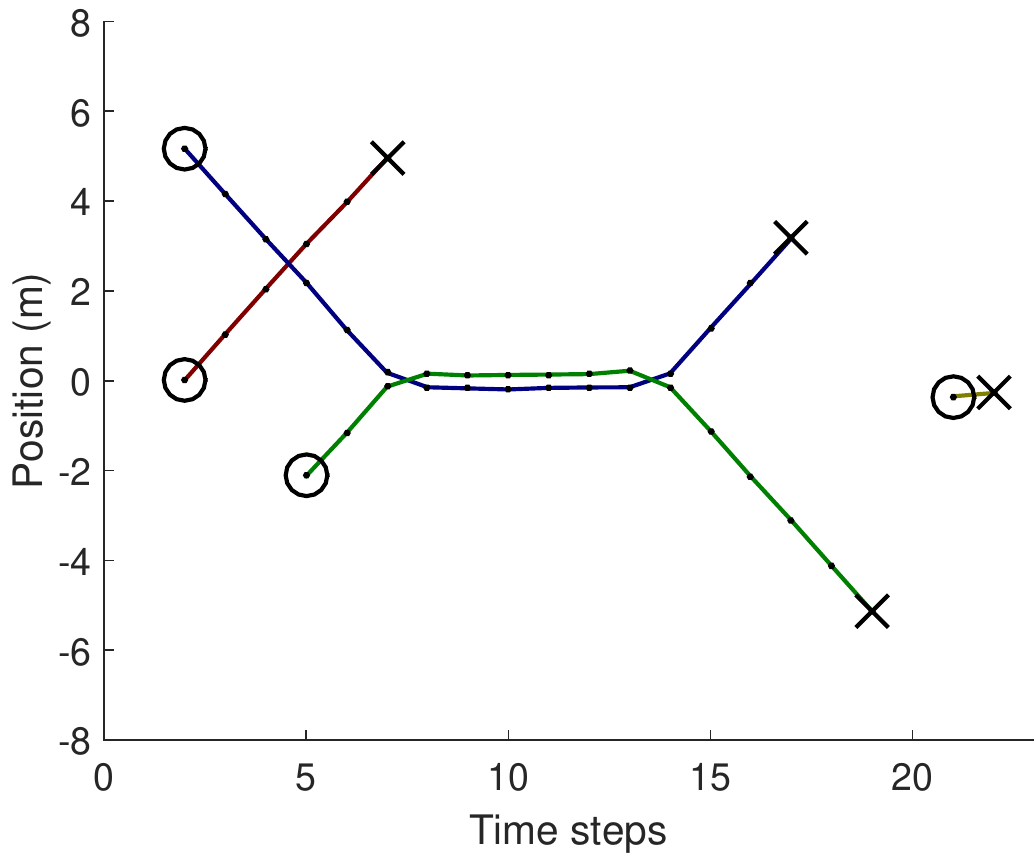} \subcaption{$Y$ with localized states, a missed target and a false track} \label{fig:case5} 
		\end{minipage}& 
		\begin{minipage}{0.3\textwidth} 
			\includegraphics[scale = 0.3]{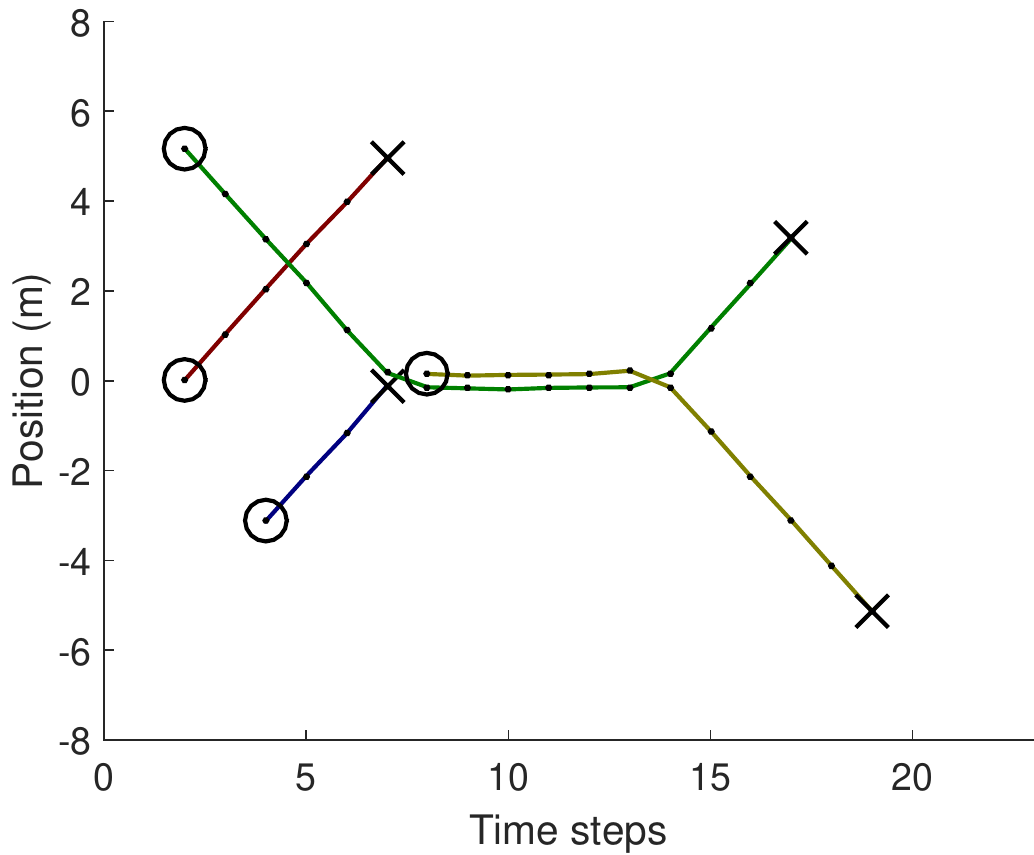}  \subcaption{$Y$ with localized states and two half switches} \label{fig:case6}
		\end{minipage}
	\end{tabular}
	\caption{Possible hypotheses for the ground truth in Figure~\ref{fig:case3Truth} based on the measurements in Figure~\ref{fig:meas}.}
	\label{fig:LPmetricIllus}
	\vspace{-0.2cm}
\end{figure*}

\begin{table*}[!b]
	\caption{LP metric $\dSh$ and Bento's metrics $d_{\text{B}}^{1}$ and  $d_{\text{B}}^{\text{comp}}$ between the estimates in Figure~\ref{fig:LPmetricIllus} and the ground truth in Figure~\ref{fig:case3Truth} for varying $c$ and $\gamma$.}
	\begin{center}
		\newcolumntype{C}{>{\centering\arraybackslash} m{0.45cm}}
		\setlength{\extrarowheight}{0.5mm}
		\begin{tabular}{|l|| C | C | C || C | C | C || C | C | C || C | C | C ||}
			\hline 
			& \multicolumn{6}{|c||}{$c=5$} & \multicolumn{6}{c||}{$c=10$} \\
			\cline{2-13}
			\multirow{2}{*}{Figure} &\multicolumn{3}{c||}{$\gamma=5$} &  \multicolumn{3}{c||}{$\gamma=10$}  & \multicolumn{3}{c||}{$\gamma=5$} & \multicolumn{3}{c||}{$\gamma=10$}  \\
			\cline{2-13}
			&$\dSh$ & $d_{\text{B}}^{\text{1}}$ & $d_{\text{B}}^{\text{comp}}$ & $\dSh$ & $d_{\text{B}}^{\text{1}}$ & $d_{\text{B}}^{\text{comp}}$ & $\dSh$ & $d_{\text{B}}^{\text{1}}$ & $d_{\text{B}}^{\text{comp}}$ & $\dSh$ & $d_{\text{B}}^{\text{1}}$ & $d_{\text{B}}^{\text{comp}}$  \\
			\hline 
			~\ref{fig:case1} &  8.2  &   8.2  &  8.2  &  8.2  &  8.2 &  8.2  &  8.2  &  8.2  &  8.2  &   8.2  &  8.2  &  8.2  \\
			~\ref{fig:case2} & 13.2 & 13.2 & 13.2 & 13.2 & 13.2 & 13.2 & 18.2 & 18.2 & 18.2 & 18.2 & 18.2 & 18.2 \\
			~\ref{fig:case3} & 21.7 & 16.7 & 21.7 & 31.7 & 21.7 & 31.7 & 26.7 & 21.7 & 26.7 & 36.7 & 26.7 & 36.7 \\
			~\ref{fig:case4} & 16.7 & 11.7 & 16.7 & 26.7 & 16.7 & 26.7 & 16.7 & 11.7 & 26.7 & 26.7 & 16.7 & 26.7 \\
			~\ref{fig:case5} & 15.8 & 15.8 & 15.8 & 15.8 & 15.8 & 15.8 & 23.3 & 23.3 & 23.3 & 23.3 & 23.3 & 23.3 \\
			~\ref{fig:case6} & 13.5 & 13.5 & 18.5 & 18.5 & 18.5 & 27.2 & 13.5 & 13.5 & 18.5 & 18.5 & 18.5 & 28.5 \\
			\hline 
		\end{tabular}
	\end{center}
	\label{tab:res}
	\vspace{-0.8cm}
\end{table*}

Here, we illustrate the behaviour of the LP metric, given by Proposition~\ref{def:LPreMetric}, for varying  values of $c$ and $\gamma$ using an MTT example. We also compare the values returned by the LP metric to Bento's metrics with switching cost based on 1-norm and component-wise 1-norm.  We have set $p=1$ and $\dbase$ as Euclidean distance in the metric for these simulations.

We consider a multiple target tracking scenario, where we use the notation, models and the Bayesian closed form solution for sets of trajectories in \cite{garcia2015setsTraj}. We consider a target state $x\in \R^{2}$ that consists of one-dimensional position and velocity for ease of illustration. The targets can be born from $2$ similar single-target densities, 
\begin{equation}
\beta_{1}(x)= \beta_{2}(x)=\mathcal{N}\Bigg(x; \begin{bmatrix} 0 \\0 \end{bmatrix}, \begin{bmatrix} 25 & 0 \\ 0 & 1 \end{bmatrix} \Bigg). \nonumber
\end{equation}
The probabilities that there are $0$, $1$ (from either of the densities) or $2$ new born targets at each time are $0.85$, $0.1$ and $0.05$, respectively. The probability for a target to survive to the next time instant is $0.9$, and the corresponding state is governed by the state transition density 
\begin{equation}
	g(x^{k}|x^{k-1}) = \mathcal{N}\Bigg(x^{k}; \begin{bmatrix} 1 & 1 \\ 0 & 1\end{bmatrix} x^{k-1}, \frac{1}{10}\begin{bmatrix} 1/3 & 1/2 \\ 1/2 & 1\end{bmatrix}\Bigg).\nonumber
\end{equation}
We consider a batch duration of $T= 22$. 

We consider the standard measurement model \cite{mahler2007statistical}. We obtain positional measurements of the targets from the sensors with probability $p_{D} = 0.95$ with the target measurements  generated according to $\mathcal{N}\Big(\cdot;\begin{bmatrix} 1 & 0 \end{bmatrix} x^{k}, 10^{-3}\Big)$. We observe Poisson clutter, which is uniformly distributed in the interval $(-10, \quad 10)$ and there is an average of $1$ clutter measurement per scan.

The position components of the targets in the ground truth and the observed measurements are shown in Figures~\ref{fig:case3Truth} and Figure~\ref{fig:meas}, respectively. For the considered models, the filtering multi-trajectory density, on the RFS of trajectories, is a multi-Bernoulli mixture \cite{garcia2015setsTraj}, which can be expanded so that the probability of existence of each Bernoulli component is either zero or one \cite{Angel18_b}, which is referred to as $\mathrm{MBM}_{01}$. Each mixture component of the $\mathrm{MBM}_{01}$ represents a global hypothesis. The filtering recursion to obtain the multi-trajectory density in $\mathrm{MBM}_{01}$ form is given in \cite{garcia2015setsTraj}, and could be extended to sets of labelled trajectories without changes in the filtering recursion \cite[Sec. IV.A]{garcia2015setsTraj}. The starting times, deaths and mean positions for the trajectories of some of the global hypotheses are shown in Figure~\ref{fig:LPmetricIllus}. We have considered these hypotheses as we think they are insightful to illustrate the behaviour of the metric. 

We first analyze the LP metric between the ground truth and the posterior mean of these hypotheses. We want to point out that the LP metric and the multi-dimensional assignment metric, computed by the Viterbi algorithm, returned the same values for these scenarios. The results are presented in Table~\ref{tab:res}. As can be seen from the tables and the figures, for fixed $c$, when the switching cost parameter $\gamma$ is increased, the metric values increase for the cases with full or half switches in Figures~\ref{fig:case3}, ~\ref{fig:case4}, and~\ref{fig:case6}. Similarly, for fixed $\gamma$, when $c$ is increased, the metric values increase for the cases with missed and/or false targets in Figures~\ref{fig:case2}, ~\ref{fig:case3} and ~\ref{fig:case5}. For the case in Figure~\ref{fig:case3} which has two track switches and a false track, the metric value increases for increase in both $c$ and $\gamma$. It can also be observed that the case in Figure~\ref{fig:case1} is always returned as the most accurate one irrespective of the parameters choice, which agrees with intuition.

Table~\ref{tab:res} also contains the results for Bento's metric with 1-norm and component-wise 1-norm denoted $d_{\text{B}}^{\text{1}}$ and $d_{\text{B}}^{\text{comp}}$, respectively. We recall that these metrics correspond to the LP relaxation of the B1 and B2 metrics explained in Section \ref{subsec:motiv}. It can be immediately observed that for the cases in Figures~\ref{fig:case1}, ~\ref{fig:case2} and ~\ref{fig:case5} which have no track switches, the values are identical between the Bento's metrics and the proposed LP metric. 

There are differences in the metric values for the scenarios in Figure~\ref{fig:case3}, ~\ref{fig:case4} and ~\ref{fig:case6}. For the case in Figure~\ref{fig:case6}, there is one track switch (or two half switches from the perspective of $\bY$) between time steps $7$ and $8$. This switch contributes $\gamma$ as the switching cost to the proposed LP metric, $\dSh$ and Bento's metric with 1-norm $d_{\text{B}}^{1}$, thus resulting in identical metric values. However, $d_{\text{B}}^{\text{comp}}$ also includes the switches in the $\ast$-trajectories, similar to the case in Figure~\ref{fig:egcompnorm_b}, and therefore has a higher switching cost leading to a higher metric value. 

For the cases in Figures~\ref{fig:case3} and ~\ref{fig:case4}, there are two track switches between time steps $11$ and $12$, which implies a switching cost of $2\gamma$ in the proposed metric $\dSh$. This value also matches with $d_{\text{B}}^{\text{comp}}$, as the switches are only between the real trajectories and it counts all the switches. However, in $d_{\text{B}}^{1}$, the switching cost contribution is $\gamma$ regardless of the number of switches (if non-zero) at a particular time, which is not the expected behaviour.

\section{Conclusion}\label{sec:concl}
In this paper, we have proposed a multi-dimensional assignment metric that quantifies the distance between two sets of trajectories. This metric captures the localization error for properly detected targets, missed targets, false targets and track switches. The penalty over the track switches is based on the changes in the sequence of assignments/unassignments. This characteristic leads to the concept of half switches, which avoids the addition of $\ast$-trajectories and the penalty of track switches that involve $\ast$-trajectories.

When the number of trajectories is small, the metric can be computed using the Viterbi algorithm. For larger problems, we have proposed a bound to the multi-dimensional assignment metric, which is also a metric and can be computed using linear programming. For all the results presented in the paper, the linear programming metric is identical to the multi-dimensional assignment metric. We have also extended these metrics to the space of random sets of trajectories, which is useful to performance evaluation and optimal estimator design.

While we think our metric captures the main penalties associated to multiple target tracking estimates in a mathematically consistent and intuitive way, metric design for sets of trajectories is an open area of research. Different metrics could be designed for different types of applications. For example, new metrics could be designed based on Bento's metrics, generalising Bento's metrics or completely new approaches. It may also be possible to generalise both Bento's families of distances and metrics to also include our metric, e.g., by extending the switching cost to be dependent on trajectories indices, and treat switches between different indices differently, as suggested in \cite{bento2016metric}.

\bibliographystyle{IEEEtran}
\bibliography{setTrajMetricPaper}
\clearpage 

\begin{center}
	{\LARGE{}Supplementary material of "A metric on the space of finite sets of trajectories for evaluation of multi-target tracking algorithms"}
	\par\end{center}{\LARGE \par}

\appendices
\section{}\label{sec:Append_Bento_metric}

In this appendix, we show that the proposed metric does not belong to Bento's family of metrics. Bento's family of metrics only includes switching costs  $\mathcal{K}\left(\cdot\right)$ that are bi-invariant on permutations \cite[Chap. 6]{Diaconis_book88}. This implies that any reordering of the indices of the elements in $\bX$ and in $\bY$ does not change the value of the metric, and we therefore have a function for sets of trajectories.   

In addition, we should note that not all members of Bento's family of distances are functions over sets of trajectories, as this family does not set constraints on the switching cost, which can make a distance depend on the indexing on the elements of the set. If we constrain this family of distances to have bi-invariant switching cost, to define functions on sets of trajectories, then the following proof also shows that our metric does not belong to Bento's family of distances. 

We should first note that two metrics are identical if and only if they take the same values for all sets of trajectories. Unless the localization and missed/false target costs are identical, for the same assignments/permutations, it is easy to construct examples with trajectories of length 1 where the metrics take different values. To ensure identical localization and missed/false target costs for equivalent assignments we can choose $p=1$ and $c=2M$, $M$ as defined in \cite{bento2016metric}, and use the same base metric $d_{b}\left(\cdot,\cdot\right)$. 

In this section, we will consider cases in which localization costs dominate the switching costs in the metric computation. That is, the entries of the assignments/permutations that correspond to existing trajectories in both sets are decided based on the locations of the trajectories at that time step. In these examples, the difference (apart from $p$) can only arise in the possible value of the switching cost, as the localization and missed/false target costs are fixed and equivalent in Bento's families and in the proposed metric, with the above choices of parameters.

We consider the three scenarios in Figure \ref{fig:c_eg2}, in which localization costs dominate switching costs, where our metric returns three different switches: Scenario a) has
two switches (switching cost $2\gamma$), b) one switch (cost $\gamma$)
and c) 1.5 switches (cost $1.5\gamma$), considering that localization
costs dominate switching cost, e.g., $\gamma\ll c$.

%%%%%%%%%%%%%%%%%% Metric of different type examples %%%%%%%%%%%%%%%%%%%%%%%%%
\tikzset{x=0.8cm,y=0.8cm, every text node part/.style={align=center}, every node/.style={font=\scriptsize, inner sep=1pt,outer sep=0pt, minimum size=2pt,  draw}}
\begin{figure}[th!]
\begin{center}
\begin{minipage}[t]{0.14\textwidth}
\centering
\begin{tikzpicture}
	\def \lenx {2}; 	\def \leny {2};\def \del{0.5};\def \ylim {-0.5};
	\def \DEL {2};
	% X1
	\node [rectangle, label = below left:$X_{1}$](x11) at (1, 0){};
	\foreach \x in {2,..., \lenx} {
		\node [rectangle](x1\x) at (\x, 0){};}
	\foreach \x in {2,..., \lenx} {
		\pgfmathtruncatemacro{\cur}{\x}
		\pgfmathtruncatemacro{\next}{\x - 1}
	\draw (x1\cur)--(x1\next);}
	
	% X2
	\node [rectangle, label = above left:$X_{2}$](x21) at (1, \del+\DEL){};
    \foreach \x in {2,..., \lenx} {
		\node [rectangle](x2\x) at (\x, \del+\DEL){};}
	\foreach \x in {2,..., \lenx} {
		\pgfmathtruncatemacro{\cur}{\x}
		\pgfmathtruncatemacro{\next}{\x - 1}
	\draw (x2\cur)--(x2\next);}
	
	% Y1
	\node [circle, label = above left:$Y_{1}$](y11) at (1, \del){};
	\foreach \x in {2,..., \leny} {
		\node [circle](y1\x) at (\x, \del){};}
		
	% Y2
	\node [circle, label = below left:$Y_{2}$](y21) at (1, \DEL){};
	\foreach \x in {2,..., \leny} {
		\node [circle](y2\x) at (\x, \DEL){};}
	
    % Y1 Y2 swap
    \draw [densely dotted](y11)--(y22);
    \draw [densely dotted](y12)--(y21);

    % to mark DELTA	
	\draw [<->]([xshift=-10pt]x11.north west) -- ([xshift=-10pt]y11.south west) node[draw=none,fill=none,midway,left] {$\Delta$};
    \draw [<->]([xshift=-10pt]y21.north west) -- ([xshift=-10pt]x21.south west) node[draw=none,fill=none,midway,left] {$\Delta$};

    % to mark delta
	\draw [<->]([xshift=10pt]y12.south east) -- ([xshift=10pt]y22.north east) node[draw=none,fill=none,midway,right] {$\delta$};  
	
%	  \foreach \x in {1, ..., \lenx}{
%   \draw (\x , \ylim) -- (\x, \ylim-0.1) node[draw = none, below] {$\x$};}            
%  \draw [-](1, \ylim) -- (\lenx, \ylim) node[draw=none, right] {$k$};
\end{tikzpicture}
\subcaption{}
\label{fig:c_eg2b}
\end{minipage}
\begin{minipage}[t]{0.16\textwidth}
\centering
\begin{tikzpicture}
	\def \lenx {2}; 	\def \leny {2};\def \del{0.4};\def \ylim {-0.5};
	\def \DEL {1.4}; \def \DELD {0.8};
	% X1
	\node [rectangle, label = below left:$X_{1}$](x11) at (1, 0){};
	\foreach \x in {2,..., \lenx} {
		\node [rectangle](x1\x) at (\x, 0){};}
	\foreach \x in {2,..., \lenx} {
		\pgfmathtruncatemacro{\cur}{\x}
		\pgfmathtruncatemacro{\next}{\x - 1}
	\draw (x1\cur)--(x1\next);}
	
	% X2
	\node [rectangle, label = above left:$X_{2}$](x21) at (1, \del+\DEL){};
    \foreach \x in {2,..., \lenx} {
		\node [rectangle](x2\x) at (\x, \del+\DEL){};}
	\foreach \x in {2,..., \lenx} {
		\pgfmathtruncatemacro{\cur}{\x}
		\pgfmathtruncatemacro{\next}{\x - 1}
	\draw (x2\cur)--(x2\next);}
	
	% X3
	\node [rectangle, label = above left:$X_{3}$](x31) at (1, \del+\DEL+\DELD){};
    \foreach \x in {2,..., \lenx} {
		\node [rectangle](x3\x) at (\x, \del+\DEL+\DELD){};}
	\foreach \x in {2,..., \lenx} {
		\pgfmathtruncatemacro{\cur}{\x}
		\pgfmathtruncatemacro{\next}{\x - 1}
	\draw (x3\cur)--(x3\next);}	
		
	% Y1
	\node [circle, label = above left:$Y_{1}$](y11) at (1, \del){};
	\node [circle](y12) at (2, \DEL){};	
	\draw [densely dotted](y11)--(y12);
		
    % to mark DELTA	    
    \node [draw=none](dmy) at (1,\DEL){};
	\draw [<->]([xshift=-10pt]x11.north west) -- ([xshift=-10pt]y11.south west) node[draw=none,fill=none,midway,left] {$\Delta$};
    \draw [<->]([xshift=-10pt]dmy.north west) -- ([xshift=-10pt]x21.south west) node[draw=none,fill=none,midway,left] {$\Delta$};

    % to mark delta
    \node [draw=none](dmy) at (2,\del){};
	\draw [<->]([xshift=10pt]y12.south east) -- ([xshift=10pt]dmy.north east) node[draw=none,fill=none,midway,right] {$\delta$};  
	
    % to mark delta2	
	\draw [<->]([xshift=10pt]x32.south east) -- ([xshift=10pt]x22.north east) node[draw=none,fill=none,midway,right] {$\delta_2$};  	
	
%	  \foreach \x in {1, ..., \lenx}{
%    \draw (\x , \ylim) -- (\x, \ylim-0.1) node[draw = none, below] {$\x$};}            
%  \draw [-](1, \ylim) -- (\lenx, \ylim) node[draw=none, right] {$k$};
\end{tikzpicture}
\subcaption{}
\label{fig:c_eg2a}
\end{minipage}
\begin{minipage}[t]{0.16\textwidth}
\centering
\begin{tikzpicture}
	\def \lenx {2}; 	\def \leny {2};\def \del{0.5};\def \ylim {-0.5};
	\def \DEL {2};
	% X1
	\node [rectangle, label = below left:$X_{1}$](x11) at (1, 0){};
	\foreach \x in {2,..., \lenx} {
		\node [rectangle](x1\x) at (\x, 0){};}
	\foreach \x in {2,..., \lenx} {
		\pgfmathtruncatemacro{\cur}{\x}
		\pgfmathtruncatemacro{\next}{\x - 1}
	\draw (x1\cur)--(x1\next);}
	
	% X2
	\node [rectangle, label = above left:$X_{2}$](x21) at (1, \del+\DEL){};
    \foreach \x in {2,..., \lenx} {
		\node [rectangle](x2\x) at (\x, \del+\DEL){};}
	\foreach \x in {2,..., \lenx} {
		\pgfmathtruncatemacro{\cur}{\x}
		\pgfmathtruncatemacro{\next}{\x - 1}
	\draw (x2\cur)--(x2\next);}
	
	% Y1
	\node [circle, label = above left:$Y_{1}$](y11) at (1, \del){};
		
	% Y2
	\node [circle, label = below left:$Y_{2}$](y21) at (1, \DEL){};
	\foreach \x in {2,..., \leny} {
		\node [circle](y2\x) at (\x, \DEL){};}
	
    % Y1 Y2 swap
    \draw [densely dotted](y11)--(y22);

    % to mark DELTA	
	\draw [<->]([xshift=-10pt]x11.north west) -- ([xshift=-10pt]y11.south west) node[draw=none,fill=none,midway,left] {$\Delta$};
    \draw [<->]([xshift=-10pt]y21.north west) -- ([xshift=-10pt]x21.south west) node[draw=none,fill=none,midway,left] {$\Delta$};

    % to mark delta
    \node [draw=none](dmy) at (2,\del){};
	\draw [<->]([xshift=10pt]dmy.south east) -- ([xshift=10pt]y22.north east) node[draw=none,fill=none,midway,right] {$\delta$};  
	
%    \foreach \x in {1, ..., \lenx}{
%    \draw (\x , \ylim) -- (\x, \ylim-0.1) node[draw = none, below] {$\x$};}            
%  \draw [-](1, \ylim) -- (\lenx, \ylim) node[draw=none, right] {$k$};
\end{tikzpicture}
\subcaption{}
\label{fig:c_eg2c}
\end{minipage}
\caption{Scenarios for the counterexample in Appendix \ref{sec:Append_Bento_metric}. We consider  $\delta\gg c$,  $\delta_2\gg c$ and $\Delta=0$, which implies that localization errors are zero. }
\label{fig:c_eg2}
\end{center}
\vspace{-0.5cm}
\end{figure}

Given two sets of trajectories $\mathbf{X}$ and $\mathbf{Y}$, the corresponding sequence of permutations in Bento's distances is denoted as $\Sigma=\left(\sigma\left(1\right),...,\sigma\left(T\right)\right)$
and each permutation is of length $\left|\mathbf{X}\right|+\left|\mathbf{Y}\right|$ \cite{bento2016metric}. We will show that no permutations $\sigma\left(1\right),$ $\sigma\left(2\right)$ in Bento's family of metrics can provide these switching
errors with the same localization and missed and false target costs. As  $\mathcal{K}\left(\cdot\right)$ is bi-invariant in Bento's metrics, without
loss of generality, we can consider that $\sigma\left(1\right)=[1,2,3,4]$
and analyse the output for different values of $\sigma\left(2\right)$. 

\subsubsection{Figure \ref{fig:c_eg2} (a)}

Only $\sigma\left(2\right)=[2,1,3,4]$ and $\sigma\left(2\right)=[2,1,4,3]$
produce a zero localization cost, as the proposed metric. Therefore,
so that the minimum over these permutations agrees with the switching
cost of the proposed metric, we must have that
\begin{align*}
\min\left[\mathcal{K}\left(\sigma\left(1\right),[2,1,3,4]\right),\mathcal{K}\left(\sigma\left(1\right),[2,1,4,3]\right)\right] & =2\gamma,
\end{align*}
where $\sigma\left(1\right)=[1,2,3,4]$.

\subsubsection{Figure \ref{fig:c_eg2} (b)}

Our metric produces 0 localization costs, 4 costs associated to missed
targets and one track switch. In order to have the same costs in Bento's metrics for
localization and missed targets, we must have that the second element
in $\sigma\left(2\right)$ is 1. In addition, to obtain the switching
cost $\gamma$, the first element in $\sigma\left(2\right)$ must
be different from 2 since we know from Example a) that the switching
costs for $\sigma\left(2\right)=[2,1,3,4]$ and $\sigma\left(2\right)=[2,1,4,3]$
are both higher or equal than $2\gamma$. The other possible values
for $\sigma\left(2\right)$ all produce the same localization/missed
costs, and the minimum switching cost among all those values for $\sigma\left(2\right)$
must therefore be $\gamma$:
\begin{align*}
& \min\left[\mathcal{K}\left(\sigma\left(1\right),[3,1,2,4\right),\mathcal{K}\left(\sigma\left(1\right),[3,1,4,2\right)\right.\\
& \:\left.\mathcal{K}\left(\sigma\left(1\right),[4,1,2,3\right),\mathcal{K}\left(\sigma\left(1\right),[4,1,3,2\right)\right]=\gamma,
\end{align*}
where $\sigma\left(1\right)=[1,2,3,4]$.
\subsubsection{Figure \ref{fig:c_eg2} (c)}

The proposed metric produces 0 localization costs, 1 missed target
cost, and 1.5 switches. In order to have the same
costs for localization, missed and false targets with Bento's metrics, the second element
in $\sigma\left(2\right)$ must be 1. In addition, as in Example b),
the first element in $\sigma\left(2\right)$ cannot be 2, as this
case was covered in Example a) and the switching cost was higher or
equal than $2\gamma$. Then, the rest of possible values of $\sigma\left(2\right)$
produce the same localization cost, and therefore the minimum over
the switching costs must be $1.5\gamma$ so that it agrees with our
metric
\begin{align*}
& \min\left[\mathcal{K}\left(\sigma\left(1\right),[3,1,2,4\right),\mathcal{K}\left(\sigma\left(1\right),[3,1,4,2\right)\right.\\
& \:\left.\mathcal{K}\left(\sigma\left(1\right),[4,1,2,3\right),\mathcal{K}\left(\sigma\left(1\right),[4,1,3,2\right)\right]=1.5\gamma,
\end{align*}
where $\sigma\left(1\right)=[1,2,3,4]$. In Examples (b) and (c), we
have the same minimization problem for Bento's metrics but it must
produce two different outputs to agree with our metric. As this is
not possible, this shows that our metric cannot be recovered by Bento's
metrics. The proof is also valid for Bento's distances with a bi-invariant $\mathcal{K}\left(\cdot\right)$, which ensures that the family of distances are defined over sets of trajectories.

\section{}\label{sec:Append_metric}
In this appendix, we first prove that $\dSh(\cdot, \cdot)$ in \eqref{eq:LPreMetric} is a metric in Section \ref{subsec:Append_metric}. Then, in Section \ref{subsec:LP_proof}, we prove that $\dSh(\cdot, \cdot)$ in \eqref{eq:LPreMetric} can be computed using LP.

\subsection{Proof of metric properties}\label{subsec:Append_metric}
The non-negativity, identity and symmetry properties of the metric in \eqref{eq:LPreMetric} are immediate from the definition. Below we prove the triangle inequality. The proof in this section is done for the LP metric, where the optimization is over $W^{k}\in \hW_{\bX, \bY}$. The proof is analogous for the multi-dimensional assignment metric in \eqref{eq:binLP}, where the optimisation is over $W^{k} \in \W_{\bX, \bY} \subset\hW_{\bX, \bY}$, and therefore also holds for the multi-dimensional assignment metric $\dS(\bX, \bY)$, see Lemma \ref{lem:mdBinLp}.

We denote the objective function in \eqref{eq:LPreMetric} as $\dSh(\bX,\bY, W^{1:T})$ as a function of the $W$ matrices. 
The outline of the proof is as follows: We assume that we have three sets of trajectories $\bX$, $\bY$ and $\bZ$.  Let $W_{\bX, \bY}^{\star} \in \hW_{\bX, \bY}$, $W_{\bX, \bZ}^{\star} \in \hW_{\bX, \bZ}$ and  $W_{\bZ, \bY}^{\star}\in \hW_{\bZ, \bY}$ be the weight matrices that minimize $\dSh(\bX,\bY, W_{\bX, \bY}^{1:T})$, $\dSh(\bX,\bZ, W_{\bX, \bZ}^{1:T})$ and $\dSh(\bZ,\bY, W_{\bZ, \bY}^{1:T})$ respectively. We construct a matrix $W_{\bX, \bY} \in \hW_{\bX, \bY}$ from $W_{\bX, \bZ}^{\star} \in \hW_{\bX, \bZ}$ and  $W_{\bZ, \bY}^{\star}\in \hW_{\bZ, \bY}$ as
\begin{flalign}
&W_{\bX,\bY}^{k}(i,j)&\nn\\
& = \begin{cases}  
1- \sum\limits_{j=1}^{\ny} W_{\bX,\bY}^{k}(i,j) & \begin{array}{l}i=1,\ldots, \nx, \: j=\ny+1\end{array}\\
1- \sum\limits_{i=1}^{\nx} W_{\bX,\bY}^{k}(i,j) & \begin{array}{l}i=\nx+1, \: j=1, \ldots, \ny\end{array}\\
0 & \begin{array}{l}i = \nx+1, \: j = \ny+1 \end{array}\\
\sum\limits_{l=1}^{\nz} W_{\bX,\bZ}^{\star k}(i,l)  W_{\bZ,\bY}^{\star k}(l,j) & \begin{array}{l} \text{otherwise.} \end{array}\\
\end{cases}& \label{eq:wmatConst}
\end{flalign}
and show that 
\begin{equation}
\dSh(\bX,\bY, W_{\bX, \bY}^{1:T})  \leq  \dSh(\bX,\bZ) + \dSh(\bZ,\bY).\label{eq:triIneq1}
\end{equation}
Combining the above result with the fact that $\dSh(\bX,\bY) \leq \dSh(\bX,\bY, W_{\bX, \bY}^{1:T})$, we get the triangle inequality 
\begin{flalign}
\dSh(\bX,\bY) & \leq \dSh(\bX,\bZ) + \dSh(\bZ,\bY).
\end{flalign}
To prove \eqref{eq:triIneq1}, we show that for any $W_{\bX, \bZ} \in \hW_{\bX, \bZ}$ and  $W_{\bZ, \bY}\in \hW_{\bZ, \bY}$, and $W_{\bX, \bY} \in \hW_{\bX, \bY}$  constructed according to \eqref{eq:wmatConst}, the following result holds:
\begin{flalign}
&\dSh(\bX,\bY, W_{\bX, \bY}^{1:T})  &\nn\\
&\qquad \leq \dSh(\bX,\bZ, W_{\bX, \bZ}^{1:T}) + \dSh(\bZ,\bY, W_{\bZ, \bY}^{1:T}).
\label{eq:triIneq}
\end{flalign}
To show that \eqref{eq:triIneq} holds, we show two separate inequalities for the switching and the localization cost using $W_{\bX, \bY}$ in \eqref{eq:wmatConst} and we bring them together towards the end. 

\subsubsection{Switching cost inequality}
For the switching cost, we show that 
\begin{flalign}
&\sum_{i=1}^{\nx} \sum_{j=1}^{\ny} |W_{\bX,\bY}^{k}(i,j) - W_{\bX,\bY}^{k+1}(i,j)|  & \nn\\
&\leq \sum_{i=1}^{\nx} \sum_{l=1}^{\nz} \bigg|W_{\bX,\bZ}^{k}(i,l) -  W_{\bX,\bZ}^{k+1}(i,l)\bigg|&\nn\\
& \quad +  \sum_{j=1}^{\ny} \sum_{l=1}^{\nz}\bigg|W_{\bZ,\bY}^{k}(l,j) - W_{\bZ,\bY}^{k+1}(l,j)\bigg| .&\label{eq:swiTri}
\end{flalign}

Starting with the left-hand side of \eqref{eq:swiTri},
\begin{flalign}
&\sum_{i=1}^{\nx} \sum_{j=1}^{\ny} |W_{\bX,\bY}^{k}(i,j) - W_{\bX,\bY}^{k+1}(i,j)|  & \nn\\
&= \sum_{i=1}^{\nx} \sum_{j=1}^{\ny} \bigg|\sum_{l=1}^{\nz} \bigg(W_{\bX,\bZ}^{k}(i,l)  W_{\bZ,\bY}^{k}(l,j) &\nn\\
&\quad -  W_{\bX,\bZ}^{k+1}(i,l)  W_{\bZ,\bY}^{k+1}(l,j)\bigg)\bigg|& \\
&\leq  \sum_{i=1}^{\nx} \sum_{j=1}^{\ny} \sum_{l=1}^{\nz} \bigg|W_{\bX,\bZ}^{k}(i,l)  W_{\bZ,\bY}^{k}(l,j) &\nn\\
&\quad -  W_{\bX,\bZ}^{k+1}(i,l)  W_{\bZ,\bY}^{k+1}(l,j)\bigg|. &
\end{flalign}
For the above inequality, we have used the inequality of the absolute value norm: $\big|\sum_{l}a_{l}\big| \leq \sum_{l}\big|a_{l}\big|$. Then, we can write

\begin{flalign}
&\sum_{i=1}^{\nx} \sum_{j=1}^{\ny} |W_{\bX,\bY}^{k}(i,j) - W_{\bX,\bY}^{k+1}(i,j)|  & \nn\\
&\leq  \sum_{i=1}^{\nx} \sum_{j=1}^{\ny} \sum_{l=1}^{\nz} \Bigg[\bigg|W_{\bX,\bZ}^{k}(i,l) -  W_{\bX,\bZ}^{k+1}(i,l)\bigg| &\nn\\
& \quad \times\frac{\big( W_{\bZ,\bY}^{k}(l,j)+  W_{\bZ,\bY}^{k+1}(l,j)\big)}{2}&\nn\\
& \quad + \bigg|W_{\bZ,\bY}^{k}(l,j) -  W_{\bZ,\bY}^{k+1}(l,j)\bigg| \frac{\big(W_{\bX,\bZ}^{k}(i,l)  +W_{\bX,\bZ}^{k+1}(i,l)\big)}{2}\Bigg].& \label{eq:swiInter1}
\end{flalign}
For the proof of the inequality in \eqref{eq:swiInter1}, notice that for $ a_{1},a_{2}, b_{1}, b_{2} \geq 0$,
\begin{flalign}
&|a_{1}a_{2}-b_{1}b_{2}| = \frac{1}{2}\bigg|(a_{1}-b_{1})(a_{2}+b_{2}) + (a_{1}+b_{1}) (a_{2}-b_{2})\bigg|&\nn\\
&\leq \big|a_{1}-b_{1}\big|\frac{(a_{2}+b_{2})}{2} + \big|a_{2}-b_{2}\big| \frac{(a_{1}+b_{1})}{2}.&
\end{flalign}
Note that in \eqref{eq:swiInter1}, using $\sum_{j=1}^{\ny}\frac{\big( W_{\bZ,\bY}^{k}(l,j)+  W_{\bZ,\bY}^{k+1}(l,j)\big)}{2}\leq 1$ and $\sum_{i=1}^{\nx} \frac{\big(W_{\bX,\bZ}^{k}(i,l)+W_{\bX,\bZ}^{k+1}(i,l)\big)}{2} \leq 1$, we get the result in \eqref{eq:swiTri}.

\subsubsection{Localization cost inequality}
First, we show two intermediate results:
\begin{flalign}
W_{\bX,\bY}^{k}(i, \ny+1)&= W_{\bX,\bZ}^{k}(i, \nz+1) & \nn\\
&  + \sum_{l=1}^{\nz}W_{\bX,\bZ}^{k}(i,l)W_{\bZ,\bY}^{k}(l, \ny+1),&\label{eq:winy1}\\
 W_{\bX,\bY}^{k}(\nx+1, j)&  = W_{\bZ,\bY}^{k}(\nz+1, j) &\nn\\
&  + \sum_{l=1}^{\nz}W_{\bX,\bZ}^{k}(\nx+1,l)W_{\bZ,\bY}^{k}(l, j).&\label{eq:wnx1j}
\end{flalign}

We prove below that the difference between the right-hand and left-hand sides of \eqref{eq:winy1} is zero.
\begin{flalign}
&W_{\bX,\bZ}^{k}(i, \nz+1) +\sum_{l=1}^{\nz}W_{\bX,\bZ}^{k}(i,l)W_{\bZ,\bY}^{k}(l, \ny+1)&\nn\\
& \qquad - W_{\bX,\bY}^{k}(i, \ny+1) &\nn\\
& = W_{\bX,\bZ}^{k}(i, \nz+1) +\sum_{l=1}^{\nz}W_{\bX,\bZ}^{k}(i,l)W_{\bZ,\bY}^{k}(l, \ny+1)&\nn\\
& \qquad  - \bigg( 1 - \sum_{j=1}^{\ny} W_{\bX,\bY}^{k}(i, j)\bigg)& \\
& = W_{\bX,\bZ}^{k}(i, \nz+1) +\sum_{l=1}^{\nz}W_{\bX,\bZ}^{k}(i,l)W_{\bZ,\bY}^{k}(l, \ny+1) &\nn\\
& \qquad - \bigg( 1 - \sum_{j=1}^{\ny} \sum_{l=1}^{\nz} W_{\bX,\bZ}^{k}(i, l)W_{\bZ,\bY}^{k}(l,j)\bigg)& \\
& = W_{\bX,\bZ}^{k}(i, \nz+1) + \sum_{l=1}^{\nz} \sum_{j=1}^{\ny+1} W_{\bX,\bZ}^{k}(i, l)W_{\bZ,\bY}^{k}(l,j) - 1& \nn\\
& = W_{\bX,\bZ}^{k}(i, \nz+1) + \sum_{l=1}^{\nz} W_{\bX,\bZ}^{k}(i, l)\sum_{j=1}^{\ny+1} W_{\bZ,\bY}^{k}(l,j) - 1& \nn\\
& = W_{\bX,\bZ}^{k}(i, \nz+1) + \sum_{l=1}^{\nz} W_{\bX,\bZ}^{k}(i, l) - 1& \\
& = \sum_{l=1}^{\nz+1} W_{\bX,\bZ}^{k}(i, l) - 1 = 0.& 
\end{flalign}
Similar proof holds for \eqref{eq:wnx1j} as well. 

We use \eqref{eq:winy1} and \eqref{eq:wnx1j} in the below derivation of the localization cost.
\begin{flalign}
&\tr\big[\big(D_{\bX,\bY}^{k}\big)^{\dagger}W_{x, y}^{k}\big] = \sum_{i=1}^{\nx+1} \sum_{j=1}^{\ny+1} D_{\bX,\bY}^{k}(i,j)W_{\bX,\bY}^{k}(i,j)&\nn\\
&=  \sum_{i=1}^{\nx} \sum_{j=1}^{\ny} D_{\bX,\bY}^{k}(i,j)W_{\bX,\bY}^{k}(i,j) &\nn\\
&+ \sum_{j=1}^{\ny} D_{\bX,\bY}^{k}(\nx+1,j)W_{\bX,\bY}^{k}(\nx+1,j)\nn & \\
&+ \sum_{i=1}^{\nx} D_{\bX,\bY}^{k}(i,\ny+1)W_{\bX,\bY}^{k}(i,\ny+1)&\nn\\
&+  D_{\bX,\bY}^{k}(\nx+1,\ny+1)\underbrace{W_{\bX,\bY}^{k}(\nx+1,\ny+1)}_{=0} &\\
& = \sum_{i=1}^{\nx} \sum_{j=1}^{\ny} D_{\bX,\bY}^{k}(i,j) \sum_{l=1}^{\nz}W_{\bX,\bZ}^{k}(i,l)W_{\bZ,\bY}^{k}(l,j) &\nn\\
& + \sum_{j=1}^{\ny} D_{\bX,\bY}^{k}(\nx+1,j) W_{\bZ,\bY}^{k}(\nz+1, j) &\nn\\
&+ \sum_{j=1}^{\ny} D_{\bX,\bY}^{k}(\nx+1,j)\sum_{l=1}^{\nz}W_{\bX,\bZ}^{k}(\nx+1,l)W_{\bZ,\bY}^{k}(l, j)&\nn\\
& + \sum_{i=1}^{\nx} D_{\bX,\bY}^{k}(i,\ny+1) W_{\bX,\bZ}^{k}(i, \nz+1)&\nn\\
& + \sum_{i=1}^{\nx} D_{\bX,\bY}^{k}(i,\ny+1) \sum_{l=1}^{\nz}W_{\bX,\bZ}^{k}(i,l)W_{\bZ,\bY}^{k}(l, \ny+1).& 
\end{flalign}

We substitute the values for $D_{\bX,\bY}^{k}(i,j)$ in the first, third and the fifth summation terms as in \eqref{eq:DMatBase} and use the triangle inequality of the base metric $d_{\mathbf{X},\mathbf{Y}}^{k}\left(i,j\right)=	D_{\mathbf{X},\mathbf{Y}}^{k}(i,j)^{1/p}$. For the second and fourth summation, we use the equalities: $D_{\bX,\bY}^{k}(\nx+1, j)=D_{\bZ,\bY}^{k}(\nz+1, j)$ and $D_{\bX,\bY}^{k}(i,\ny+1) = D_{\bX,\bZ}^{k}(i, \nz+1)$. Then,
\begin{flalign}
&\tr\big[\big(D_{\bX,\bY}^{k}\big)^{\dagger}W_{\bX,\bY}^{k}\big]&\nn\\
\leq &  \sum_{i=1}^{\nx} \sum_{j=1}^{\ny} \sum_{l=1}^{\nz}\big(d_{\bX,\bZ}^{k}(i,l)+d_{\bZ,\bY}^{k}(l,j)\big)^{p} W_{\bX,\bZ}^{k}(i,l)W_{\bZ,\bY}^{k}(l,j) &\nn\\
& + \sum_{j=1}^{\ny} D_{\bZ,\bY}^{k}(\nz+1,j) W_{\bZ,\bY}^{k}(\nz+1, j) &\nn\\
&+ \sum_{j=1}^{\ny} \sum_{l=1}^{\nz}\big(d_{\bX,\bZ}^{k}(\nx+1,l)+d_{\bZ,\bY}^{k}(l,j)\big)^{p} &\nn\\
& \quad \times W_{\bX,\bZ}^{k}(\nx+1,l)W_{\bZ,\bY}^{k}(l, j)&\nn\\
& + \sum_{i=1}^{\nx} D_{\bX,\bZ}^{k}(i,\nz+1) W_{\bX,\bZ}^{k}(i, \nz+1) &\nn\\
&+ \sum_{i=1}^{\nx} \sum_{l=1}^{\nz}\big(d_{\bX,\bZ}^{k}(i,l)+d_{\bZ,\bY}^{k}(l,\ny+1)\big)^{p}&\nn\\
& \quad \times W_{\bX,\bZ}^{k}(i,l)W_{\bZ,\bY}^{k}(l, \ny+1).&\label{eq:locTri}
\end{flalign}

We observe that the right-hand side has the form
\begin{flalign}
&\tr\big[\big(D_{\bX,\bY}^{k}\big)^{\dagger}W_{\bX,\bY}^{k}\big] \leq \sum_{i,l,j} (a_{i,l} + b_{l,j})^{p} + \sum_{j}(0 + b_{j})^{p}&\nn\\
& \: + \sum_{l,j}(a_{l} + b_{l,j})^{p} + \sum_{i}(a_{i}+0)^{p} + \sum_{i,l}(a_{i,l}+b_{l})^{p}.&\label{eq:locCostMink}
\end{flalign}
 This structure of the localization cost inequality simplifies the triangle inequality proof when we apply the Minkowski inequality. Note that we have not included the range of the indexes for notational clarify.

\subsubsection{Proof for \eqref{eq:triIneq}}
Using \eqref{eq:locCostMink} and \eqref{eq:swiTri}, we show the following result for the objective function in the overall LP cost in \eqref{eq:LPreMetric}
\begin{flalign}
&\dSh(\bX,\bY, W_{\bX, \bY}^{1:T})&\nn\\
&\leq \Bigg(\sum_{k=1}^{T} \sum_{i,l,j} (a_{i,l} + b_{l,j})^{p} + \sum_{j}(0 + b_{j})^{p} + \sum_{l,j}(a_{l} + b_{l,j})^{p}&\nn\\
& + \sum_{i}(a_{i}+0)^{p} + \sum_{i,l}(a_{i,l}+b_{l})^{p} &\nn\\
& + \sum_{k=1}^{T-1}\sum_{i=1}^{\nx} \sum_{l=1}^{\nz} \underbrace{\frac{\gamma^{p}}{2} \bigg|W_{\bX,\bZ}^{k}(i,l) -  W_{\bX,\bZ}^{k+1}(i,l)\bigg|}_{a_{n}+0} &\nn\\
& + \sum_{k=1}^{T-1}\sum_{j=1}^{\ny} \sum_{l=1}^{\nz}\underbrace{\frac{\gamma^{p}}{2}\bigg|W_{\bZ,\bY}^{k}(l,j) - W_{\bZ,\bY}^{k+1}(l,j)\bigg|}_{0+b_{n}}\Bigg)^{\frac{1}{p}}.&
\end{flalign}

Now, we use the Minkowski inequality \cite[pp. 165]{kubrusly2011elements}: $\bigg(\sum\limits_{m} \big[a_{m}+b_{m}\big]^{p}\bigg)^{\frac{1}{p}}\leq \bigg(\sum\limits_{m} a_{m}^{p}\bigg)^{\frac{1}{p}}+\bigg(\sum\limits_{m}b_{m}^{p}\bigg)^{\frac{1}{p}}$ for $p\geq 1$ and $a_{m}, b_{m}\geq 0$. Note that in the above inequality, we have several $a_{m}$'s and $b_{m}$'s that are $0$.
\begin{flalign}
&\dSh(\bX,\bY, W_{\bX, \bY}^{1:T})&\nn\\
&\leq \Bigg(\sum_{k=1}^{T} \bigg[\sum_{i,l,j} a_{i,l}^{p} + \sum_{j}0^{p} + \sum_{l,j}a_{l}^{p} + \sum_{i}a_{i}^{p} + \sum_{i,l}a_{i,l}^{p} \bigg]&\nn\\
& + \sum_{k=1}^{T-1}\sum_{i=1}^{\nx} \sum_{l=1}^{\nz} \frac{\gamma^{p}}{2} \bigg|W_{\bX,\bZ}^{k}(i,l) -  W_{\bX,\bZ}^{k+1}(i,l)\bigg|\Bigg)^{\frac{1}{p}} &\nn\\
&+\Bigg(\sum_{k=1}^{T} \bigg[\sum_{i,l,j} b_{l,j}^{p} + \sum_{j}b_{j}^{p} + \sum_{l,j}b_{l,j}^{p} + \sum_{i} 0^{p} + \sum_{i,l}b_{l}^{p}\bigg] &\nn\\
& + \sum_{k=1}^{T-1}\sum_{j=1}^{\ny} \sum_{l=1}^{\nz}\frac{\gamma^{p}}{2}\bigg|W_{\bZ,\bY}^{k}(l,j) - W_{\bZ,\bY}^{k+1}(l,j)\bigg|\Bigg)^{\frac{1}{p}}.& \label{eq:triPrlas}
\end{flalign}

Let us revisit \eqref{eq:locTri} to simplify $\sum a^{p}$ and $\sum b^{p}$ in the above terms.
\begin{flalign}
&\sum_{i,l,j} a_{i,l}^{p} + \sum_{j}0^{p} + \sum_{l,j}a_{l}^{p} + \sum_{i}a_{i}^{p} + \sum_{i,l}a_{i,l}^{p}&\nn\\
=&  \sum_{i=1}^{\nx} \sum_{j=1}^{\ny} \sum_{l=1}^{\nz}d_{\bX,\bZ}^{k}(i,l)^{p} W_{\bX,\bZ}^{k}(i,l)W_{\bZ,\bY}^{k}(l,j)&\nn\\
& + \sum_{j=1}^{\ny} \sum_{l=1}^{\nz}d_{\bX,\bZ}^{k}(\nx+1,l)^{p} W_{\bX,\bZ}^{k}(\nx+1,l)W_{\bZ,\bY}^{k}(l, j)&\nn\\
& + \sum_{i=1}^{\nx} D_{\bX,\bZ}^{k}(i,\nz+1) W_{\bX,\bZ}^{k}(i, \nz+1)&\nn\\
& + \sum_{i=1}^{\nx} \sum_{l=1}^{\nz}d_{\bX,\bZ}^{k}(i,l)^{p}W_{\bX,\bZ}^{k}(i,l)W_{\bZ,\bY}^{k}(l, \ny+1).&
\end{flalign}

Combining the first and last summations and using $\sum_{j=1}^{\ny+1}W_{\bZ,\bY}^{k}(l,j)=1$ and using $\sum_{j=1}^{\ny}W_{\bZ,\bY}^{k}(l,j)\leq1$ in the second summation, we get
\begin{flalign}
&\sum_{i,l,j} a_{i,l}^{p} + \sum_{j}0^{p} + \sum_{l,j}a_{l}^{p} + \sum_{i}a_{i}^{p} + \sum_{i,l}a_{i,l}^{p}&\nn\\
\leq&  \sum_{i=1}^{\nx} \sum_{l=1}^{\nz}d_{\bX,\bZ}^{k}(i,l)^{p} W_{\bX,\bZ}^{k}(i,l) &\nn\\
& + \sum_{l=1}^{\nz}d_{\bX,\bZ}^{k}(\nx+1,l)^{p} W_{\bX,\bZ}^{k}(\nx+1,l)&\nn\\
& + \sum_{i=1}^{\nx} D_{\bX,\bZ}^{k}(i,\nz+1) W_{\bX,\bZ}^{k}(i, \nz+1)&\\
=& \tr\big[\big(D_{\bX,\bZ}^{k}\big)^{\dagger}W_{\bX,\bZ}^{k}\big].&
\end{flalign}

Similarly we can show that $\sum_{i,l,j} b_{l,j}^{p} + \sum_{j}b_{j}^{p} + \sum_{l,j}b_{l,j}^{p} + \sum_{i} 0^{p} + \sum_{i,l}b_{l}^{p} \leq \tr\big[\big(D_{\bZ,\bY}^{k}\big)^{\dagger}W_{\bZ,\bY}^{k}\big]$. Substituting these values in \eqref{eq:triPrlas}, we get \eqref{eq:triIneq}.

\subsection{Proof for computability using LP}\label{subsec:LP_proof}
The proof for the computability of the metric in \eqref{eq:LPreMetric} using LP is along the same lines as in \cite[Theorem 10]{bento2016metric}. First, note that to compute the metric in \eqref{eq:LPreMetric}, it is enough to solve the following optimization problem:
\begin{flalign}
&\underset{\substack{W^{k} \in \hW_{\bX, \bY} \\ k = 1, \ldots, T}}{\arg \min}\sum_{k=1}^{T} \tr\big[\big(D_{\bX,\bY}^{k}\big)^{\dagger}W^{k}\big]  &\nn\\
&  \qquad + \frac{\gamma^{p}}{2}\sum_{k=1}^{T-1}\sum_{i=1}^{\nx} \sum_{j=1}^{\ny} |W^{k}(i,j) - W^{k+1}(i,j)| .&
\end{flalign}
The objective function in the above problem can be written in linear form as 
\begin{flalign}
\underset{\substack{W^{k} \in \hW_{\bX, \bY} \\ k = 1, \ldots, T\\e^{1}, \ldots, e^{T-1} \in \R}}{\arg\min}\sum_{k=1}^{T} \tr\big[\big(D_{\bX,\bY}^{k}\big)^{\dagger}W^{k}\big]  + \frac{\gamma^{p}}{2}\sum_{k=1}^{T-1}e^{k} \label{eq:LPform_Weh}
\end{flalign}
by introducing variables $e^{k}\in \R$ for $k=1, \ldots, T-1$ to the optimization problem with constraints 
\begin{flalign}
e^{k} \geq \sum_{i=1}^{\nx} \sum_{j=1}^{\ny} |W^{k}(i,j) - W^{k+1}(i,j)|. \label{eq:LP_constraint_e}
\end{flalign}
Note that, except the constraints in \eqref{eq:LP_constraint_e}, all the other constraints in \eqref{eq:binConst1}, \eqref{eq:binConst2}, \eqref{eq:binConst3} and \eqref{eq:LPConst2} are linear. We can write the optimisation problem in linear form by introducing additional variables $H^{k}(i,j) \in \R$ for $k=1, \ldots, T-1$ with the constraints:
\begin{flalign}
e^{k} &\geq \sum_{i=1}^{\nx} \sum_{j=1}^{\ny} H^{k}(i,j),& \label{eq:LPconst_e}\\
H^{k}(i,j) &\geq W^{k}(i,j) - W^{k+1}(i,j), \begin{array}{l} i = 1, \dots, \nx\\ j= 1, \ldots, \ny\end{array},& \label{eq:LPcontsh1} \\
H^{k}(i,j) &\geq  W^{k+1}(i,j)- W^{k}(i,j),  \begin{array}{l} i= 1, \dots, \nx\\ j= 1, \ldots, \ny\end{array}.&\label{eq:LPcontsh2}
\end{flalign}
\section{}\label{sec:Append_RFS_metric}
In this appendix, we prove Lemma \ref{lem:metricRFS}.
The proof is analogous for $\bar{d}_{p}^{\left(c,\gamma\right)}\left(\cdot,\cdot\right)$
and $d_{p}^{\left(c,\gamma\right)}\left(\cdot,\cdot\right)$. We consider RFSs $\mathbf{X}$ and $\mathbf{Y}$ of trajectories
with joint multitrajectory density $\pi\left(\mathbf{X},\mathbf{Y}\right)$.
By the properties of the metric, we have that $\bar{d}_{p}^{\left(c,\gamma\right)}\left(\mathbf{X},\mathbf{Y}\right)^{p'}\leq\left(c^{p}T\max\left(\left|\mathbf{X}\right|,\left|\mathbf{Y}\right|\right)\right)^{p'/p}\leq\left(c^{p}T\right)^{p'/p}\left(\left|\mathbf{X}\right|^{p'/p}+\left|\mathbf{Y}\right|^{p'/p}\right)$. Then, $\mathbb{E}\left[\bar{d}_{p}^{\left(c,\gamma\right)}\left(\mathbf{X},\mathbf{Y}\right)^{p}\right]\leq\left(c^{p}T\right)^{p'/p}\left(\mathbb{E}\left[\left|\mathbf{X}\right|^{p'/p}\right]+\mathbb{E}\left[\left|\mathbf{Y}\right|^{p'/p}\right]\right)$. The right-hand side is finite since the moments $\mathbb{E}\left[\left|\mathbf{X}\right|^{p'/p}\right]$ and $\mathbb{E}\left[\left|\mathbf{Y}\right|^{p'/p}\right]$ are assumed finite, which implies that $\sqrt[p']{\mathbb{E}\left[\bar{d}_{p}^{\left(c,\gamma\right)}\left(\mathbf{X},\mathbf{Y}\right)^{p'}\right]}$ is finite.

We need to prove the following properties to show that $\sqrt[p']{\mathbb{E}\left[\bar{d}_{p}^{\left(c,\gamma\right)}\left(\mathbf{X},\mathbf{Y}\right)^{p'}\right]}$
is a metric: definiteness, non-negativity, symmetry and the triangle
inequality. The definiteness, non-negativity and symmetry properties
are observed directly from the definition. It should be noted that,
for metrics in a probability space, the definiteness between random
variables is in the almost sure sense \cite[Sec. 2.2]{rachev2013methods}.
We proceed to prove the triangle inequality.

Let us consider three RFS $\mathbf{X}$, $\mathbf{Y}$ and $\mathbf{Z}$
of trajectories with joint density $\pi\left(\mathbf{X},\mathbf{Y},\mathbf{Z}\right)$ and finite moments $\mathbb{E}\left[\left|\bX\right|^{p'/p}\right]$, $\mathbb{E}\left[\left|\bY\right|^{p'/p}\right]$, and $\mathbb{E}\left[\left|\bZ\right|^{p'/p}\right]$.
We first apply the triangle inequality for the metric on sets of trajectories
to obtain
\begin{align}
& \sqrt[p']{\mathbb{E}\left[\bar{d}_{p}^{\left(c,\gamma\right)}\left(\mathbf{X},\mathbf{Y}\right)^{p'}\right]}\nonumber \\
& \leq\sqrt[p']{\mathbb{E}\left[\left(\bar{d}_{p}^{\left(c,\gamma\right)}\left(\mathbf{X},\mathbf{Z}\right)+\bar{d}_{p}^{\left(c,\gamma\right)}\left(\mathbf{Z},\mathbf{Y}\right)\right)^{p'}\right]}.\label{eq:RFS_metric_append}
\end{align}
Let us now consider the $L^{p'}$ space of functions on three sets
of trajectories, which is
\begin{align*}
 L^{p'} & =\left\{ f:\left\Vert f\right\Vert _{p'}=\left(\int\left|f\left(\mathbf{X},\mathbf{Y},\mathbf{Z}\right)\right|^{p'}\delta\mathbf{X}\delta\mathbf{Y}\delta\mathbf{Z}\right)^{1/p'}<\infty\right\} ,
\end{align*}
where $f\left(\cdot\right)$ is a function with adequate units such
that the set integral is well-defined. Given $f,g\in L^{p'}$,
the Minkowski inequality for $L^{p'}$ spaces is \cite{Knapp_book05}
\begin{align}
\left\Vert f+g\right\Vert _{p'} & \leq\left\Vert f\right\Vert _{p'}+\left\Vert g\right\Vert _{p'}.\label{eq:Minkowski_inequality}
\end{align}

We define
\begin{align*}
f\left(\mathbf{X},\mathbf{Y},\mathbf{Z}\right) & =\bar{d}_{p}^{\left(c,\gamma\right)}\left(\mathbf{X},\mathbf{Z}\right)\pi\left(\mathbf{X},\mathbf{Y},\mathbf{Z}\right)^{1/p'}\\
g\left(\mathbf{X},\mathbf{Y},\mathbf{Z}\right) & =\bar{d}_{p}^{\left(c,\gamma\right)}\left(\mathbf{Z},\mathbf{Y}\right)\pi\left(\mathbf{X},\mathbf{Y},\mathbf{Z}\right)^{1/p'}
\end{align*}
which implies that
\begin{align*}
\left\Vert f\right\Vert _{p'} & =\sqrt[p']{\mathbb{E}\left[\bar{d}_{p}^{\left(c,\gamma\right)}\left(\mathbf{X},\mathbf{Y}\right)^{p'}\right]}\\
\left\Vert g\right\Vert _{p'} & =\sqrt[p']{\mathbb{E}\left[\bar{d}_{p}^{\left(c,\gamma\right)}\left(\mathbf{Z},\mathbf{Y}\right)^{p'}\right]}\\
\left\Vert f+g\right\Vert _{p'} & =\sqrt[p']{\mathbb{E}\left[\left(\bar{d}_{p}^{\left(c,\gamma\right)}\left(\mathbf{X},\mathbf{Z}\right)+\bar{d}_{p}^{\left(c,\gamma\right)}\left(\mathbf{Z},\mathbf{Y}\right)\right)^{p'}\right]}.
\end{align*}
Applying the Minkowski inequality (\ref{eq:Minkowski_inequality})
to (\ref{eq:RFS_metric_append}), we obtain
\begin{align*}
& \sqrt[p']{\mathbb{E}\left[\bar{d}_{p}^{\left(c,\gamma\right)}\left(\mathbf{X},\mathbf{Y}\right)^{p'}\right]}\\
& \leq\sqrt[p']{\mathbb{E}\left[\bar{d}_{p}^{\left(c,\gamma\right)}\left(\mathbf{X},\mathbf{Z}\right)^{p'}\right]}+\sqrt[p]{\mathbb{E}\left[\bar{d}_{p}^{\left(c,\gamma\right)}\left(\mathbf{Z},\mathbf{Y}\right)^{p'}\right]}.
\end{align*}
This completes the proof of the triangle inequality and of Lemma \ref{lem:metricRFS}.

\end{document}